# AS-XAI: Self-supervised Automatic Semantic Interpretation for CNN

Changqi Sun[1,2], Hao Xu[3], Yuntian Chen[4, *], and Dongxiao Zhang[2, 4, *]

[1] School of Environmental Science and Engineering, Southern University of Science and Technology, Shenzhen 518055, Guangdong, P. R. China

[2] Department of Mathematics and Theories, Peng Cheng Laboratory, Shenzhen 518000, Guangdong, P. R. China

[3] BIC-ESAT, ERE, and SKLTCS, College of Engineering, Peking University, Beijing 100871, P. R. China

[4] Ningbo Institute of Digital Twin, Eastern Institute of Technology, Ningbo, Zhejiang 315200, P. R. China

[*] Corresponding author.

E-mail address: 12231107@mail.sustech.edu.cn (C. Sun); 390260267@pku.edu.cn (H. Xu); ychen@eitech.edu.cn (Y. Chen); dzhang@eitech.edu.cn (D. Zhang)

## Abstract

Explainable artificial intelligence (XAI) aims to develop transparent explanatory approaches for "black-box" deep learning models. However, it remains difficult for existing methods to achieve the trade-off of the three key criteria in interpretability, namely, reliability, causality, and usability, which hinder their practical applications. In this paper, we propose a self-supervised automatic semantic interpretable explainable artificial intelligence (AS-XAI) framework, which utilizes transparent orthogonal embedding semantic extraction spaces and row-centered principal component analysis (PCA) for global semantic interpretation of model decisions in the absence of human interference, without additional computational costs. In addition, the invariance of filter feature high-rank decomposition is used to evaluate model sensitivity to different semantic concepts. Extensive experiments demonstrate that robust and orthogonal semantic spaces can be automatically extracted by AS-XAI, providing more effective global interpretability for convolutional neural networks (CNNs) and generating human-comprehensible explanations. The proposed approach offers broad fine-grained extensible practical applications, including shared semantic interpretation under out-of-distribution (OOD) categories, auxiliary explanations for species that are challenging to distinguish, and classification explanations from various perspectives.

## 1. Introduction

With rapid advancements of technology and abundance of data, convolutional neural networks (CNNs)[1] have achieved remarkable performance in diversified fields, particularly in computer vision[2] and natural language processing.[3] Despite their high accuracy and broad application, deep neural network classifiers are often regarded as "black boxes", making it difficult for humans to infer general rules between input features and output results. Such black-box models do not provide insights into the transparency of the decision-making process,[4] leaving them limited in safety-sensitive applications like autonomous driving[5] or areas with significant societal impact, such as drug discovery.[6] Therefore, in the algorithm's decision-making process, the black-box model is expected to possess knowledge, judgment, and a sense of responsibility. To address this challenge, explainable artificial intelligence (XAI)[7] has attracted increasing attention, which explains the model's decision-making process and even contributes to other scientific discoveries, such as in neuroscience[8] and physics.[9]

Interpretability is defined as "the degree to which a human can understand the cause of a decision".[10] When a black-box model achieves high accuracy on classification and detection tasks, a logical question is whether the reasons for its decisions can be trusted without understanding the model. Based on past empirical reasoning and intuition, humans can infer that local concepts are parts of the same whole.[11] We expect neural networks to convey interpretable information comprehensively, similar to humans, through concepts, semantics, and vocabulary.[12] By doing so, it is possible to intuitively identify the hierarchical relationship between samples, as well as how the model understands and builds the topology between samples. Indeed, interpretability assists humans in validating model decisions and enhancing trust in models, even without additional information.

Therefore, interpretability of the black-box deep neural networks has constituted an active area of research over the past few years given the widespread adoption of deep neural networks for various tasks. There have been two mainstream methods for improving the interpretability of neural networks, including providing local XAI and global XAI for existing neural networks. We review the advantages and disadvantages of the mainstream approaches, as detailed in Extended Data Table 1. Despite the advancements made in these methods, they still fail to achieve three trade-offs in interpretability,[14] including reliability, causality, and usability. Specifically, reliability ensures that the generated explanations align with the model's internal decision-making process, causality ensures that the explanations are comprehensible, and usability ensures that the explanations are generated at a reasonable computational cost.

Local XAI focuses on why the model produced a specific decision output on a given input sample. A large body of literature has focused on perturbation-based forward propagation approaches,[15] for example, Shapley additive explanations (SHAP)[17] usually only perturbs a single test sample input or intermediate layer to calculate the impact on the output. However, repeated perturbations lead to high computational costs, and the interpretation results are challenging to understand intuitively through visualization. Another is backpropagation-based approaches,[18] in which layer-wise relevance propagation (LRP)[20] propagates the importance signal from the output neuron to the input layer through each layer of the network to explain the model output. These methods, however, are subject to different transmission rules, which will lead to dissimilar interpretation results. Saliency-based methods[21] are the most common form of neural network *post hoc* interpretation, and the most typical integrated gradients (IG)[23] calculates gradient integrals on multiple sampling points in the model input space and serves as explanations for the model's predictions. A significant problem with saliency maps is that they usually highlight the edges in the image without considering the class. As a consequence, giving highly similar interpretations for multi-classification tasks, which are costly to interpret, and the gradient shattering problem common to the model, will have an unstable impact on the explanation. Some investigations also focus on gradient-weighted class activation mapping (CAM)[24] that visualizes the attribution map of local areas of the output features by calculating the attribute mapping relationship from the input space to the output space. However, this method can only explain a local area for a specific single sample, and cannot accurately elucidate the internal reasoning process of the model.

Interpretable representations learning[26] explains the decision-making process by learning clear semantic representations of the network training process. Concept whitening (CW)[28] aligns the representations learned by the model with known concepts through the concept whitening module. Nevertheless, the latent concept space may be related to many distinct predefined concepts, which also affects the interpretation performance of the CW module. Emerging works also exist that attempt to create intrinsically interpretable models.[29] For example, ProtoPNet[30] uses a concept bottleneck structure to recognize concepts based on the distance between features and concepts, and TesNet[31] and Deformable ProtoPNet[32] adaptively capture meaningful object features. However, these types of methods only consider the impact of a specific perceptual domain on CNN decision-making during the explanation process, and the causal relationship of the entire local explanation process is unclear, resulting in limited reliability of these explanation methods. In addition, the above local XAI can only provide specific interpretations for a single sample, thus constraining the usability of the local XAI method.

Global explainable artificial intelligence (XAI), on the other hand, focuses on the interaction between features or concepts learned by the model. It is typically employed to identify the general rules across a wider range of input data, and to provide an understanding of the overall model's behavior and decision-making methods. Concept relevance propagation (CRP)[33] combines local and global explanations of the 'where' and 'what' problems of a single prediction. This method, however, is extended based on the LRP method and relies on concept correlation propagation, which still requires certain computing resources. The above methods can also only predict a single sample, which constrains the usable scope of the interpretation method. To remedy this problem, semantic explainable artificial intelligence (S-XAI)[34] extracts common traits from CNNs in an explicit interpretable semantic space to provide global explanations of model decisions. However, this method relies on subjective manual annotation and many data pre-processing methods, which may be tedious and time-consuming. Manual annotation of semantics that relies on human knowledge may also result in inaccurate interpretation, and its practical application is limited. This method also ignores the sensitivity analysis of the semantic concepts of different neurons in the model, leading to an imprecise explanation of the cause-and-effect relationships in decision-making.

In order to overcome the existing limitations, this paper proposes a self-supervised automatic explainable artificial intelligence (AS-XAI). It utilizes self-supervised approaches to extract common traits from datasets in an autonomous manner, and combines with visualization methods to accurately capture global decisions of the model. A demonstration of the AS-XAI framework output is presented in **Figure 1**, which shows how AS-XAI provides global semantic interpretation for diverse images. With AS-XAI, we aim to provide a global semantic interpretation tool for deep models with the following three main contributions:

1) Automatic global semantic interpretation: A self-supervised interpretable framework is constructed to automatically extract important features without manual conceptual annotations, reducing the label bias and semantic confusion caused by human factors in feature extraction. In the process of global interpretation, the semantic probability of principal components and the importance of high-rank-sensitive semantics in filters are integrated to understand the semantic decision-making process of CNN explicitly.

2) Strictly orthogonal semantic spaces: Qualitative and quantitative evidence are presented for the enhanced orthogonality of the extracted semantic spaces in AS-XAI, which improves the validity of the semantic interpretation.

3) Fine-grained extensible interpretability: The proposed AS-XAI elucidates CNN decision-making regarding 'where, what, and why', effectively addressing various real-world tasks (see Figure 1), including fine-grained determination of interpretability of OOD categories, auxiliary interpretation of important features of species that are difficult for humans to understand, and multi-category semantic interpretation from different perspectives.

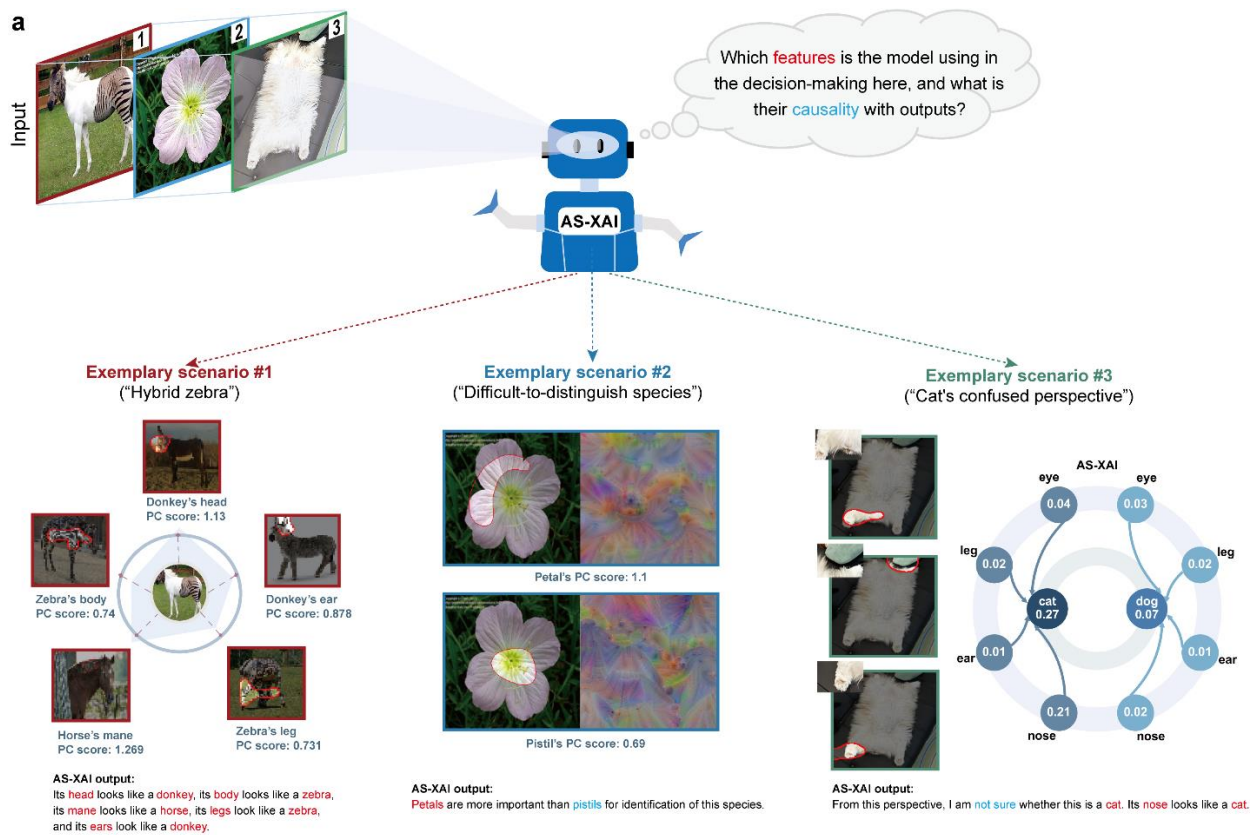

**b** AS-XAI fine-grained extensible application and its elaboration of semantic decisions "where, what, why"

**Figure 1.** AS-XAI is a novel way of using self-supervised global semantic retrieval and evaluation to grant intrinsic interpretability to CNN. a) Three exemplary inference scenarios encompass hybrid species from OOD categories, difficult-to-distinguish species, and animal images with confusing perspectives. b) The proposed AS-XAI can automatically find the common and different semantic concepts from the existing training samples and analyze the three decision problems of "where, what, and why", which explains the semantic decision of the model from a global perspective. AS-XAI demonstrates compatibility with diverse image datasets, providing fine-grained and scalable visual interpretation.

## 2. Methodology

This section demonstrates a novel approach to semantic global interpretation by AS-XAI (Section 2.1). Then, we elaborate on the method of orthogonal concept extraction (Section 2.2). Finally, the methods for interpreting semantic features are described (Section 2.3), including high-rank semantic importance analysis, row-centered PCA extraction of semantic common traits, and the visualization process of interpretation results. Experiment details and links to the open source code are publicly available on GitHub (https://github.com/qi657/AS-XAI).

### 2.1. Overview of AS-XAI

In this work, AS-XAI utilizes the commonly used CNN network as a backbone and provides semantic interpretation through the self-supervised extraction of meaningful semantic feature space from the model. Row-centered PCA[34] and semantic global similarity assessment are used to elucidate the decision-making process for the specific target. Here, take the VGG-19 network[35] as an example and add a global average pooling (GAP) layer before the fully connected (FC) layer. The GAP layer reduces the dimension of the feature map and extracts more global space-invariant features. Drawing on the TesNet[31] prototype network's design idea, two additional layers of downsample convolution are added before the classification layer for more efficient feature extraction. As shown in **Figure 2**, the global semantics interpretation is accomplished hierarchically in three steps. Figure 2a represents orthogonal concept extraction. The self-supervised orthogonalization method is employed to train Proto-CNN. This method automatically maps extracted semantic features onto the embedding space of the Grassmann manifold, obtaining local pixel-level class similarity interpretations (for detailed information, refer to Section 2.2). The semantic interpretation process is depicted in Figure 2b. The orthogonal concepts extracted from the comprehensible embedding space in Figure 2a are respectively input into the pre-training model to obtain semantic importance, visualizations, and principal component scores (PCS) to achieve comprehensive semantic interpretation. On the one hand, as illustrated in Figure 2b, singular value decomposition (SVD)[37] high-rank decomposition of the $N_f$ dimensional filters is performed in the last layer of the network, and the output of the features extracted from each filter is visualized. To obtain the importance evaluation of the network for different semantics, we calculate the rank-weighted average of the same semantics across all filters. On the other hand, common semantic features are extracted using row-centered PCA (refer to Section 2.3 for detailed information). These compressed principal components are utilized for calculating semantic probability and generating semantic feature visualizations (refer to Figure 2c). Through these processes, the ultimate semantic interpretation of specific images can be generated. Therefore, AS-XAI provides a human-understandable global semantic interpretation system with pixel-level local category interpretation and global semantic space interpretation.

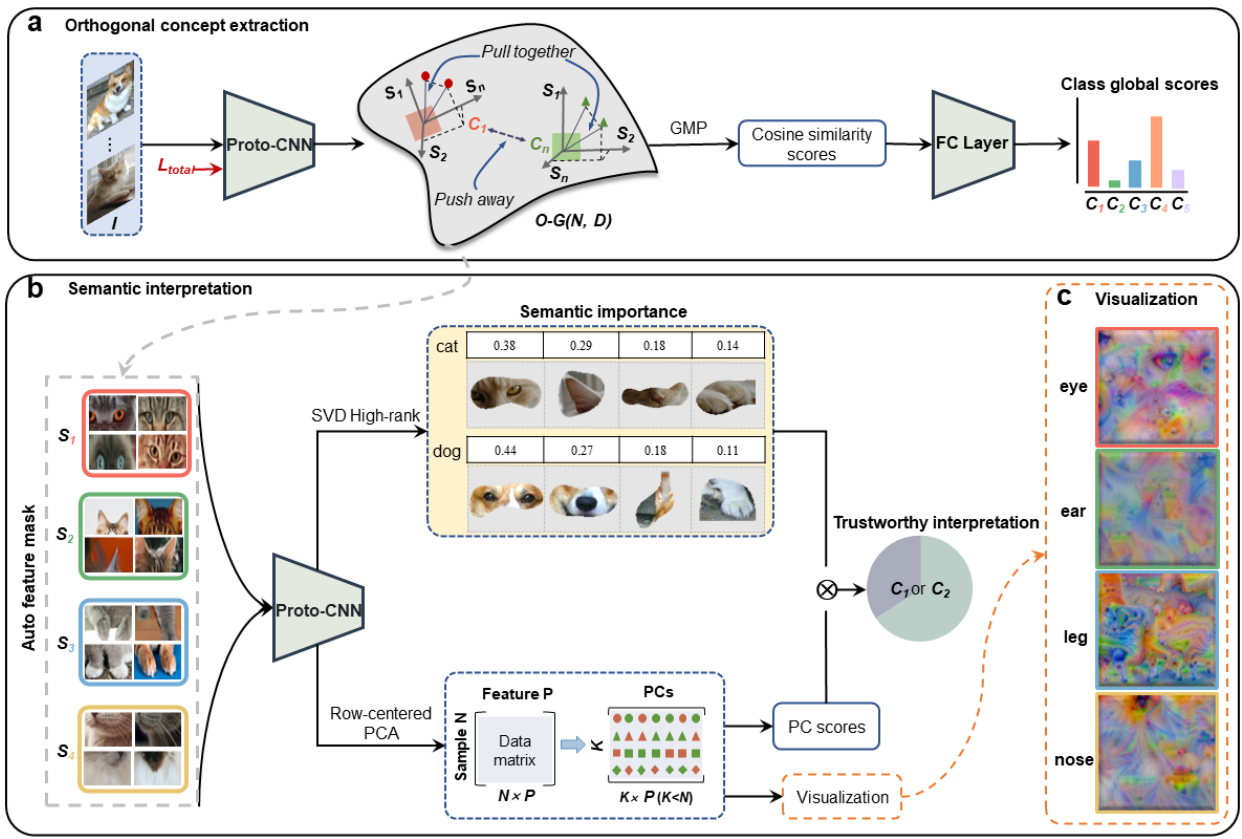

**Figure 2.** Overview of the proposed AS-XAI framework. All stages share the same backbone model. a) Orthogonal concept extraction is used for self-supervised automatic extraction of important visual concepts specific to the category, and global cosine similarity is computed for all subcategories. b) Semantic

interpretation uses the auto feature mask obtained from the comprehensible embedding space of the Grassmann manifold as the input for this stage and outputs interpretable statements specific to the semantics of the image through semantic importance assessment by SVD high-rank and semantic probability computation by row-centered PCA. c) Visualization of the common semantic space.

## 2.2. Orthogonal Concept Auto Extraction

In this work, to extract meaningful feature masks, we draw inspiration from TesNet[31] and propose Proto-CNN to self-supervise the extraction of semantic information in the category space, as described in Figure 2a. The Proto-CNN used in this paper consists of a regular convolutional neural network as the backbone. We incorporated a GAP layer and two 1×1 downsampling convolutional layers before the FC layer. The GAP layer preserves the spatial information extracted by the previous convolutional layer and facilitates feature visualization. In the additional convolutional layer, the output channel is 128, using ReLU as the activation function of the first convolutional layer, and sigmoid activates the last layer of convolution. We take the cat and dog dataset as an example, and: input the dataset into Proto-CNN; the model extracts the feature map of $bs×D×7×7$; the model maps it to the comprehensible embedding space of the Grassmannian manifold; the map initializes a base vector of $M×D×1×1$; the [0,1] normal distribution is used as the convolution kernel to calculate the cosine similarity between the feature map and the vector; and the semantic feature map of $bs×M×7×7$ is obtained. Inter-class clustering loss and inter-class separation loss are used, and the formulae are as follows：

$$\mathcal{L}_{\text{aggregation}} = \frac{1}{n}\sum_{i=1}^{n} \min_{j:a_j \in A(y_i)} \min_{P \in \text{patches}(Z_i)} -\cos\left(\theta\left(a_j^{'} \cdot P\right)\right) \tag{1}$$

$$\mathcal{L}_{\text{separation}} = \frac{1}{n}\sum_{i=1}^{n} \min_{j:a_j \notin A(y_i)} \min_{P \in \text{patches}(Z_i)} \cos\left(\theta\left(a_j^{'} \cdot P\right)\right) \tag{2}$$

where $z$ is the feature map of the image extracted by CNN; $a_j$ represents the extracted prototype feature $a_j \in A(c)$, where $c$ represents the category; $A(c)$ indicates the basis vector in the category; $a_j^{'}$ is a positive perturbation near the tiny range of the current sample feature space $a_j^{'} = a_j + \sigma\varepsilon$, in which the purpose is to make the difference between the real feature $P$ extracted by the network and the positive sample prototype. The space is closer, and the negative sample prototype is more dispersed, where $\varepsilon{\sim}N(0, I)$, $\sigma{=}0.01$. The row space orthogonality is used to enforce independence between all basis vectors within a category, $\mathcal{L}_{orth} = \sum_{c=1}^{C} \| A^{(c)}A^{(c)^{\top}} - \mathbb{I}_M \|_F^2$; and $\mathbb{I}_M$ is an $M{\times}M$ identity matrix, where $\|\cdot\|_F{}^2$ is the matrix Frobenius norm. Minimizing the correlation between basis vectors allows us to obtain diverse semantic concepts and guarantees no overlapping between concepts. The orthogonal loss constrains the different feature basis vectors within a category without semantic overlap and separates the categories well in the embedding space. To effectively distinguish between different categories within the embedding space, the category space of feature extraction is projected onto a low-dimensional manifold. The Grassmannian manifold projection metric is then employed to assess the similarities and dissimilarities of features across distinct category spaces. We separate all classes of extracted basis vector masks by maximizing each semantic space projection metric as follows:

$$\mathcal{L}_{ss} = \frac{-1}{\sqrt{2}} \sum_{c_1}^{C-1} \sum_{c_2}^{C} \left\| A^{(c_1)^{\top}} A^{(c_1)} - A^{(c_2)^{\top}} A^{(c_2)} \right\|_F \tag{3}$$

Given the training dataset $\{(x_i, y_i)\}_{i=1}^{n}$, the difference between the predicted and true labels is measured by the cross-entropy loss, which assists the model to learn effective feature representations and optimize classification accuracy:

$$L_{ce} = -\frac{1}{n} \sum_{i=1}^{n} \sum_{c=1}^{C} y_{ic} \, log \, \phi_c(x_i; \omega_{conv}, A) \tag{4}$$

where $n$ is the number of training images; $y_{ic}$ corresponds to the $c^{\text{th}}$ element of the one-hot encoded label of the sample; and $\Phi_c$ denotes the $c^{\text{th}}$ element of $\Phi$. The output layer adopts the softmax function so that $\sum_{c=1}^{C} \phi_c(x_i; \omega_{conv}, A){=}1$ and $\phi_c(x_i; \omega_{conv}, A){\geqslant}0, \forall c, i, w_{conv}, A$. For the model mapping to the streamlined space with class $c$ basis vectors $a_j, a_j \in A(c)$, we associate each basis vector with a specific image block, each of which traces back from the same class to its preselected nearest image block: $a_j \leftarrow arg \max_{p \in P_c} p^{\top} a_j$, where $P_c = \{\tilde{p}: \tilde{p} \in patches\ (Z_i) \forall i\ s.t.\ y_i = c\}$, which makes the embedding manifold space and provides a comprehensible bridge between the model's embedding space and the human-understandable domain.

Finally, the joint optimization problem in the embedding space is solved end-to-end: $L_{total} = L_{ce} + \lambda_1 L_{orth} + \lambda_2 L_{ss} + \lambda_3 L_{separation} + \lambda_4 L_{aggregation}$, where $\lambda_1$, $\lambda_2$, $\lambda_3$, and $\lambda_4$ are the hyper-parameters to balance the corresponding terms. In specific experiments, we use $\lambda_1$ as 1, $\lambda_2$ as $-1e^{-7}$, $\lambda_3$ as -0.08, and $\lambda_4$ as 0.8. We can effectively learn the embedding space of different categories to obtain better feature representation. The similarity activation map between the optimized feature and the base vector obtained by the embedding space is converted into a single cosine similarity scalar of $bs \times M$ using global max pooling (GMP), which is used to reflect the specific data of the base vector and the model. The initial connection weight of the FC layer is -0.5, and the weight range during the optimization process is [-0.5, +1.0] to represent the correct and incorrect connection weights, respectively. The global similarity score is multiplied by the corresponding weight to obtain the global cosine similarity result under all categories of a specific image.

The training process of the model includes three stages: 1) optimize the Proto-CNN by initializing the parameters, utilizing the Adam optimizer[13] to update the parameters of the additional convolutional layer, and performing $N_1$ epoch warm-up using the pre-trained CNN backbone; 2) the comprehensible embedding space self-supervised mask is implemented, updating each basis vector with the highest similarity potential training image patch. This process trains for $N_2$ epochs; and 3) convex optimization of the FC layer is performed. These three stages are alternately trained until the model converges, releasing $M$ feature masks for each category. In the experiment, we set $M$ to 100, $N_1$ to 2, and $N_2$ to 10.

### 2.3. Semantic Interpretation

#### 2.3.1. High-Rank Sensitivity of Semantic Concepts

High-rank decomposition[36] as a common network compression scheme, by the high-dimensional matrix, is represented as two or more low-dimensional matrices to approximate the convolution operation, to capture data over the network to extract after presenting the most crucial information or influence the decision of potential factors. Typical implementations include the SVD and PCA[38] or non-negative matrix factorization (NMF).[39] Previous work confirmed that the average rank of the feature maps generated by a single filter is almost unchanged.[40] By the decomposition of the two low-dimensional matrices and high-rank feature maps of the filter, the amount of information is low, in which the higher is the rank of the feature map, the less is the amount of information. In this study, we employ SVD to perform high-rank decomposition on the feature maps corresponding to all filters in the final layer of the model:

$$o_j^i(z,:,:) = \sum_{i=1}^{r} \sigma_i u_i v_i^T = \sum_{i=1}^{r'} \sigma_i u_i v_i^T + \sum_{i=r'+1}^{r} \sigma_i u_i v_i^T \tag{5}$$

where $z$ denotes the input image; $O_j^i(z, :, :)$ denotes the corresponding feature map of the input image $z$ under the $i$th filter of the $j$th convolutional layer; $r = \text{Rank}(o_j^i(z,:,:))$, where $r' < r$, and $\sigma_i$, $u_i$, and $v_i$ denote the singular values, left singular vectors, and right singular vectors, respectively; and a feature map of rank $r$ can be decomposed into a lower rank feature map of $r'$, i.e., $\sum_{i=1}^{r'} \sigma_i u_i v_i^T$ and the remaining higher-rank feature map, i.e., $\sum_{i=r'+1}^{r} \sigma_i u_i v_i^T$. A higher-rank feature map better retains the information in the feature map and has a greater impact on the model. Therefore, higher-rank feature maps can be used as a reliable measure of the importance of information in specific model layers, as shown in Figure 2b.

To evaluate the sensitivity of different filters to the extracted semantic concepts, we average the ranks corresponding to the same semantic concepts extracted by different filters in the same convolutional layer:

$$R_s(z) = \frac{1}{K_s} \sum_{k=0}^{K_s} r_s^k(z) \tag{6}$$

where $R_s(z)$ denotes the average rank of input image $z$ in the current convolutional layer when extracting a specific semantic concept; $K_s$ represents the number of filters for extracting a certain same semantic concept; and $r_s^k(z)$ indicates the rank of the input image $z$ for extracting a semantic concept under the $k$th filter.

### 2.3.2. Row-Centered PCA for Common Traits

The row-centered PCA is a feature extraction and dimensionality reduction technique applied to datasets.[34] Diverging from traditional column PCA, row-centered PCA aims to reduce the dimensionality of feature maps generated by a CNN across the sample space. This reduction is used to extract a concise set of principal components representing common semantic features learned by CNN.[41] This work utilizes row-centered PCA to distill common traits among samples from the transparently embedded semantic space, facilitating the comprehension of salient and differential features between samples (as described in Figure 2b). By entering the feature matrix $W$ of the input data and calculating the covariance matrix, the covariance matrix is horizontally reduced by using SVD. The detailed formulae are as follows:

$$\widehat{W}_{i,j} = W_{i,j} - \overline{W}_i \tag{7}$$

$$P = \frac{1}{m-1} \widehat{W} \widehat{W}^T \tag{8}$$

where $i$ is the index of rows; $j$ is the index of columns; $m$ is the total sample; and $P$ is the covariance matrix. The SVD decomposition of the $P$ matrix $S = U\Sigma V^T$ yields $U, \Sigma$, and $V \in R^{n \times n}$, where $\Sigma$ is a singular value matrix; the rank of $\Sigma$ is $r < \min\{n, m\}$; and we retain the first $k$ ($k \leqslant r$) columns of the singular value $U$ as $U_k \in R^{n \times k}$, and obtain principal component (PC) information by $W_k = W^T U_k \in R^{m \times k}$, where each column of $W_k$ is a PC, and the elements of PC are called PCS. Therefore, the information ratio of the given $i^{\text{th}}$ PC is calculated as follows:

$$\pi_i = \sum_{i=1}^{k} \lambda_i \left( \sum_{i=1}^{m} \lambda_i \right)^{-1} = \sum_{i=1}^{k} \lambda_i (\text{tr}\ (P))^{-1} \times 100\% \tag{9}$$

where $tr(P)$ denotes the trace of the covariance matrix $P$; $\lambda_i$ is the corresponding eigenvalue; and $\pi_i$ represents the percentage of retained k-dimensional PC information in the total variance.

### 2.3.3. Visualization of Semantic Space

Data analysis and visualization are the easiest ways to gain insight into candidate vectors, aiming to enable understanding of the characteristics of network learning in a human-comprehensible manner.

According to the data type and model architecture, the data corresponding to the representation of the input data and the region indicated by the candidate concept vector can be visualized.[39] In this work, common traits of global semantics in datasets extracted by self-supervised models are visualized understandably (as shown in Figure 2c), and here we describe the approach in detail. The objective of feature visualization is to address a minimization problem. In order to enhance the spatial smoothness of feature visualization images and reduce noise and other oscillations, this study uses the total variational (TV) norm to constrain the noise and detail level of the visualization results. Therefore, the minimization constraint problem is formulated as follows:

$$z^* = \arg\min_{z \in R^{C \times H \times W}} \left( \|\Phi(z) - \Phi_0\|^2 + \lambda \left( \frac{1}{CHW} \| \nabla_x \Phi(z) + \nabla_y \Phi(z) \| \right) \right) \tag{10}$$

where $\Phi_0$ is the ground truth of the target data feature; $\Phi(z)$ is the feature map with shape $C \times H \times W$ extracted by the model for the input image, where $C$ is the number of channels, and $H$ and $W$ are the height and width of the feature map, respectively; and $\lambda$ is employed to control the magnitude of regularization. TV regularization encourages the generated images to have smoother textures and edges, resulting in clearer and more realistic image visualizations.

The following stochastic gradient descent (SGD) method is used to optimize the above minimization constraint problem:

$$z^{n+1} = z^n - \beta \cdot \nabla z^* \tag{11}$$

where we initialize an image where all pixels are 0; $z^n$ is the image after $n$ iterations of the network; and $\beta$ is the learning rate. During the experiment, $\beta$ is set to 0.05, and $n$ is 4000.

### 2.3.4. Trustworthy Interpretation

In this work, more causal relationships between semantic information and prediction conclusions can be provided by the proposed AS-XAI. Trustworthy interpretation is demonstrated via a semantic probability bubble ring diagram and a global structural similarity histogram. The semantic probability bubble ring diagram intuitively indicates the semantic decision-making process of the model in identifying various images. Additionally, the global similarity histogram provides fine-grained pixel-level interpretation between categories. For the input image $z$, we first compute the cosine similarity of each image patch with the dataset's mask features in the orthogonal Grassmann manifold embedding space $O\text{-}G(N, D)$, where $N$ is the sample category; and $D$ represents the feature dimension to obtain a global category difference histogram as a basis for the fine-grained interpretation of the multiclassification.

Next, we focus on the semantic space of CNN's predicted categories. In order to measure the PCS of image z in semantic space, the semantic probability $P_s(z)$ and semantic importance $R_s(z)$ are weighted, which is written as follows:

$$P_s(z) = \frac{cdf\left(X = A_s(z)\right) - cdf\left(X = A_{min}\right)}{cdf\left(X = A_{max}\right) - cdf\left(X = A_{min}\right)} \tag{12}$$

$$PCS_s(z) = P_s(z) \times R_s(z) \tag{13}$$

where $PCS_s(z)$ is the PC score of the input image $z$ under the semantic space; $A_s(z)$ denotes the probability density distribution of the weighted average activation values of the input image $z$; $cdf$ is the cumulative distribution function fitted to a normal distribution; $A_{min}$ and $A_{max}$ are the minimum and maximum values of $A_s$ in the distribution, respectively. For each semantic concept, $\Delta_{max}PCS_s(z)$ represents the maximum difference between the predicted category and the found category in all semantic spaces; $PCS_{max}(z)$ denotes the maximum value of PCS across all semantic spaces; and $\Delta PCS(z)$ represents the difference in semantic PCS between predicted and found categories. The confidence assessment criteria in the semantic space are determined by $PCS_{max}(z)$, $\Delta PCS(z)$, and $\Delta_{max}PCS_s(z)$, and different probability intervals under different terms (e.g., definitely, probably, puzzlingly) are used to identify the confidence level. The rules for generating interpretations for AS-XAI are provided in Extended Data Table 2. All interpretable statements are automatically generated by synthesizing global category difference histograms and semantic bubble ring diagrams, and AS-XAI can provide efficient and trustworthy explanations for any input image.

## 3. Experiments

To evaluate the effectiveness of the proposed AS-XAI, in this section, the datasets and implementation details are first introduced (Section 3.1). Subsequently, we present the semantic feature space of AS-XAI self-supervised extraction (Section 3.2). Next, the sensitivity of semantic concepts is examined to understand how VGG-19 responds to different attributes (Section 3.33). We then explore the fine-grained, scalable applications of AS-XAI (Section 3.4). Finally, an orthogonal ablation experiment of the semantic space is conducted to elucidate the effect of orthogonality on interpretability (Section 3.5).

### 3.1. Datasets and Implementation Details

In this study, various datasets are utilized to evaluate the performance of AS-XAI, as listed in **Table 1**. For the multi-classification task, the cat and dog datasets were utilized, which contains the 12-classification dataset of cats from the paddle image classification learning tournament and the 70 dog breeds-images from Kaggle. For the interpretability of OOD categories, Animals with Attributes2 (AwA2)[16] are used as the training dataset, which contains 37322 images distributed in 50 animal categories. The Animals 151 dataset was used as the inference dataset. The Oxford 102 dataset[27] and the classify-leaves dataset from Kaggle were utilized in experiments to interpret species that are challenging for human understanding. The Oxford 102 dataset consists of 1764 training images and 363 test images, while the classify-leaves dataset comprises 724 training images and 35 test images.

Proto-CNN was implemented in PyTorch, and the dataset was augmented by rotation, skew, shear, and random distortion. Subsequently, the dataset was downscaled to a pixel-level resolution of $224 \times 224$. The training process of Proto-CNN employed the Adam optimizer with a learning rate initialized to $1e^{-4}$ and a batch size of 32. Training and testing were performed using an NVIDIA GeForce RTX 3090 GPU.

**Table 1.** List of datasets used in this paper.

| Dataset | Classifier |
|---|---|
| Cats and dogs | Cats/dogs |
| AwA2[16] | Wild animals species |
| Oxford 102 Flower Dataset[27] | Flowers species |
| Classify-leaves | Leaves species |
| Animal image dataset | Animal species (lions/tigers) |

### 3.2. Self-supervised Extraction of Common Semantic Spaces

The common traits of species have become an essential basis for category recognition. In previous investigations, Grad-cam[25] was usually used to specify the most obvious areas of image response, but these methods only reflected individual features and could not generate common semantic spaces. S-XAI[34] manually defines semantic concepts in datasets and labels them to obtain global semantic interpretation, which is both time-consuming and ineffective for extracting semantic information from species with irregular feature shapes. In addition, many label biases appeared in the manually extracted semantic features, affecting the validity of subsequent interpretable results.

The key to handling the above challenges is to utilize the self-supervised method to extract features, which is more computationally efficient and avoids potential biases introduced by artificial labeling. In this section, row-centered PCA is employed on the embedding space from self-supervised orthogonal clustering to abstract and visualize common traits of datasets. This assists in understanding semantic concepts and overcoming certain problems, such as semantic confusion. The results of the common semantic space for the cat and dog datasets are displayed in **Figure 3**. The proposed AS-XAI randomly selected *Ns* samples and obtained important semantic feature masks through self-supervised learning. Visualization confirmed that these masks corresponded to distinct categories, including ears, eyes, nose, and legs. By extracting the common traits represented by the 1$^{st}$ PC, it is shown that the feature space of AS-XAI, which extracts common semantics, is pure and embodies distinct semantic features (Figure 3a,c). In addition, we analyzed the correlation between feature distribution and normal distribution for all semantic spaces using quantile-quantile (q-q) plots (Figure 3b,d). The coefficient of determination ($R^2$) measured conformity to the normal distribution, with values near 1 indicating higher conformity. The q-q plot shows that $R^2$ is consistently above 0.98, which indicates a robust correlation between feature distribution and normal distribution across semantic spaces. The difference between the commonality space obtained by AS-XAI and previous studies[34] is discussed in Supplementary Information S.2, and it can be seen that many semantic overlaps occur in the semantic space by manually extracting semantic features. The quantitative comparison between our method and previous methods on q-q plots also offers obvious advantages. Combined with the subjective and objective analysis of the results, we confirmed that the common traits space of the AS-XAI self-supervised mask reflects a strong semantic pertinence.

Results show that the semantic features extracted by the AS-XAI method find a good representation from data through self-supervision, which can deepen understanding of the model's significant semantics without being influenced by human prior factors. Additionally, it eliminates label bias across various semantic concepts and reduces the time costs associated with human intervention. The method's high interpretation efficiency enables it to be transferred to larger, more complex tasks.

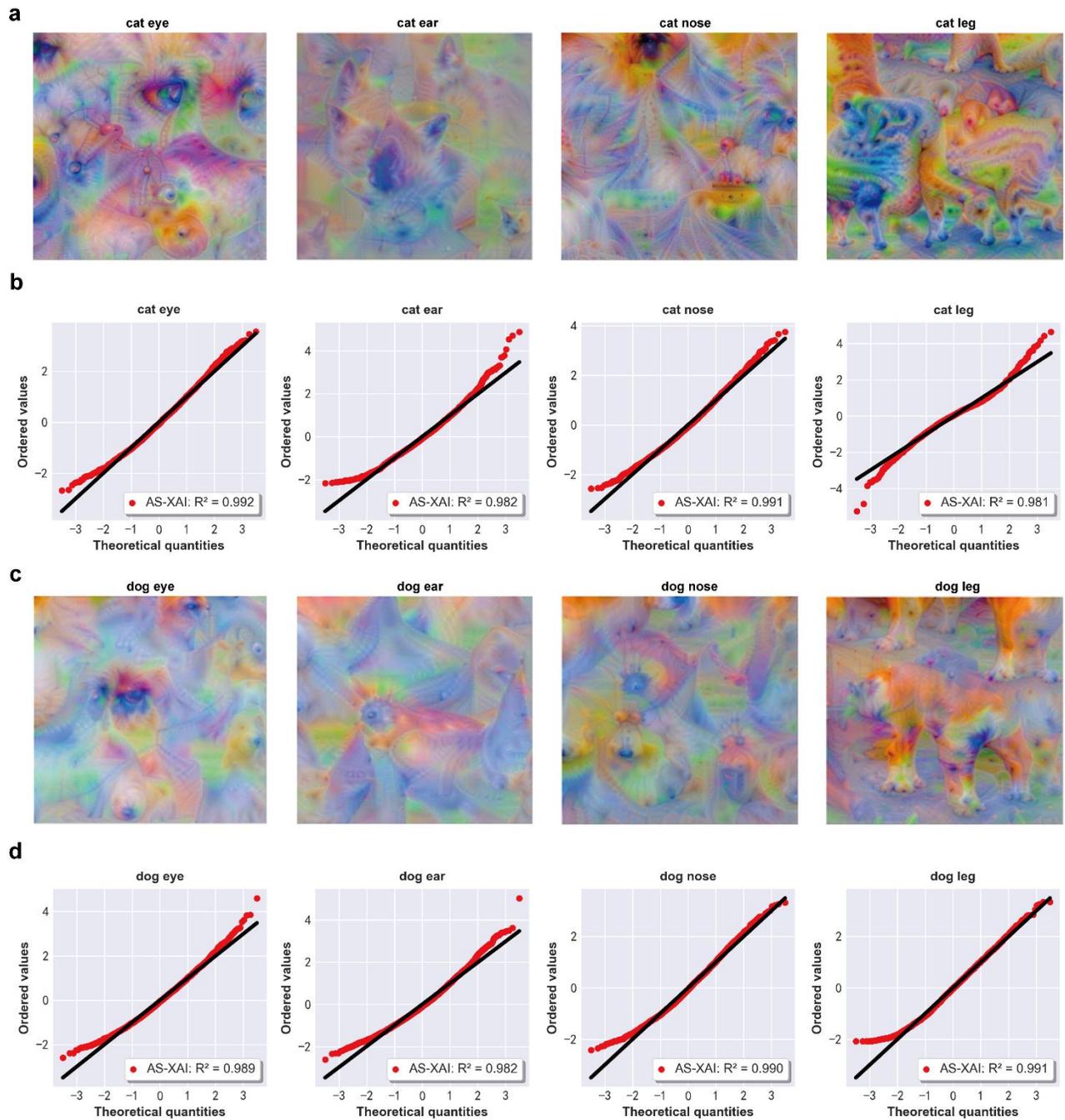

**Figure 3.** The results for self-supervised extraction of common traits. Visualization of the significant semantic spaces of cats (a) and dogs (c) in the AS-XAI from left to right for eye, nose, ear, and leg in the 1st PC ($N_s$=400). The q-q plots depict the semantic space distributions of cats (b) and dogs (d), where red dots represent the semantic distributions of AS-XAI. The black line quantifies the fitted normal distribution line.

### 3.3. Sensitivity of Semantic Concepts

There are some differences in the human visual system's understanding of the semantics of different species. From the perspective of the perceptual domain, humans' recognition of objects is more inclined to the difference of shapes. Furthermore, variability exists in the CNN's sensitivity to different semantic concepts. When recognizing cats, humans do not treat their characteristics equally, but rather distinguish them hierarchically. For instance, the facial features of cats may be more important than the body. However, previous interpretability research[31] treated the importance of extracted features equally and did not discuss the sensitivity of different features recognized by CNNs, which obscures the causality that affects interpretable results. Through embedded SVD high-rank decomposition, this work explores the cognitive differences of AS-XAI in various image semantic spaces, addressing the challenge of lacking strict causality in semantic concept interpretation without additional computational cost.

After conducting extensive experiments, we observed that data from the same category exhibit similar features on corresponding filters, and the ranking of each filter layer remains consistent. Moreover, the semantic sensitivity of different categories under the same filter varies, and more details of the experiments and results can be found in Supplementary Information S.4. The results of semantic sensitivity analysis of different data by AS-XAI on the $N_f$ filters at the last layer of the model are shown in **Figure 4**. The horizontal coordinates in the second column represent values from high-rank decomposition, while the vertical coordinate corresponds to different semantic categories. Dashed lines denote the ratio of ranks associated with distinct semantics compared to total semantic rank. The third column presents semantic concept visualizations under corresponding filters and global sensitivity scores. Higher ratios indicate the greater significance of semantics to feature extraction; whereas, lower ratios denote weaker significance. The model consistently ranked eyes highest and legs lowest in sensitivity for cat and dog samples. The results are reported in Figure 4a,b. Moreover, the task of classification of flowers is examined, in which the features are more complex and irregular. Figure 4c,d shows that AS-XAI is more sensitive to petals than pistils of two types of flowers, such as Black-eyed Susan and primrose. It is worth noting that the last feature extraction layer of VGG-19 has six filter channels for primrose flowers that have not extracted any meaningful semantic space, which implies that these filters are redundant. These two sets of experimental results demonstrate that AS-XAI exhibits varying sensitivity to recognizing different semantic aspects of species.

Therefore, the features extracted by AS-XAI for high-rank solutions can efficiently analyze the sensitivity of different neurons to extract semantic concepts, which significantly improves interpretability and clarity. We assign its weight based on semantic interpretation results, thus embodying the robust cognitive logic of AS-XAI and making the underlying factors influencing interpretable results more explicit. In addition, the experiment also reflects that the understanding of species by CNN shares certain similarities with the human cognitive system.

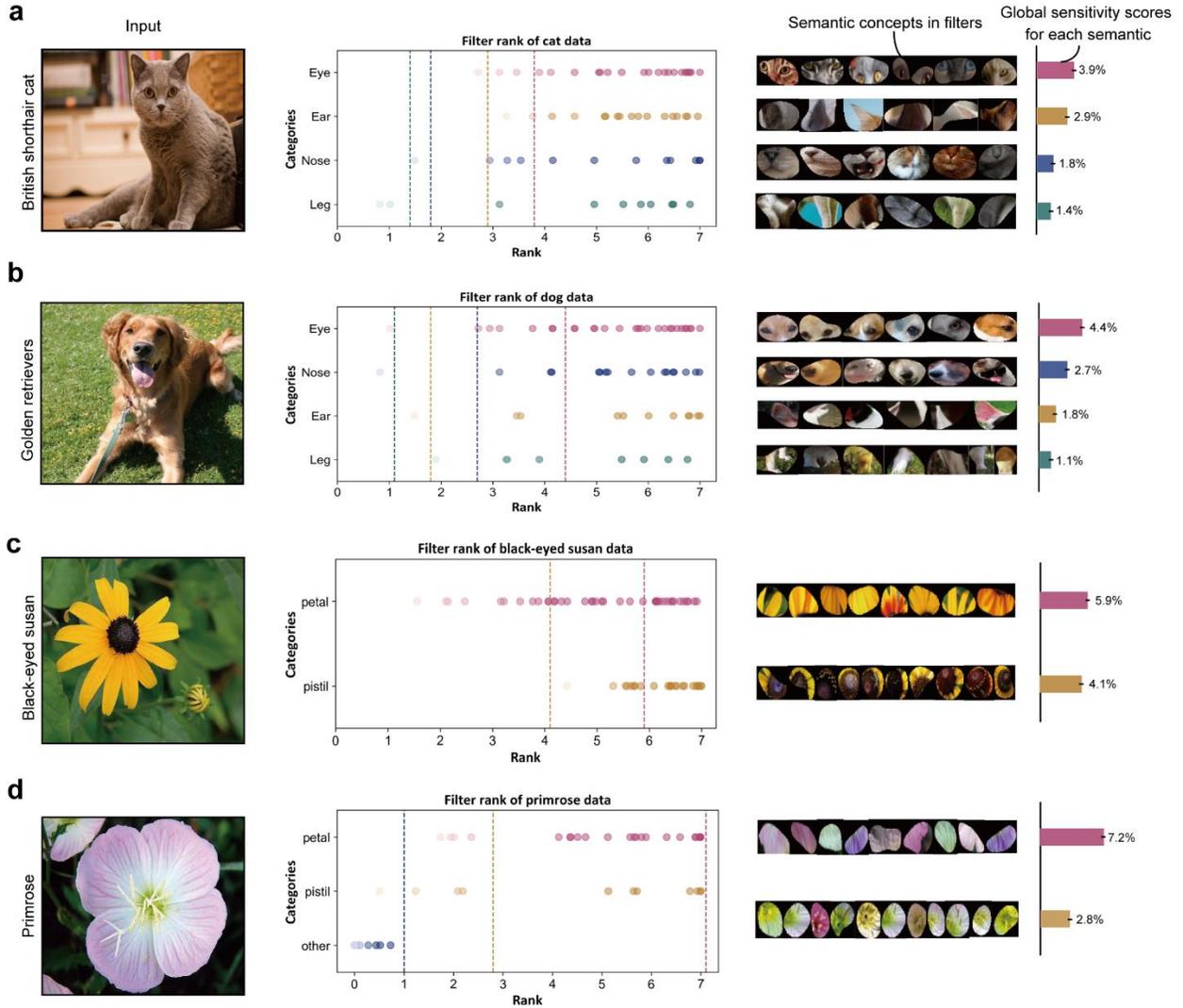

**Figure 4.** Filter rank of different data-sensitive spaces. The sensitivity of the last feature extraction layer of VGG-19 ($N_f = 64$) to different semantic spaces under the data of cat (a), dog (b), Black-eyed Susan (c), and primrose (d). Scatter points of different colors represent the semantic concepts under the filter, and dotted lines of different colors represent the mean value of the ranks of each semantic space under the total semantic space after high-rank decomposition. The filter extracted category-dependent semantic concepts, highlighting the same concepts for a given input image category, and obtained a sensitivity score for each semantic.

### 3.4. Fine-Grained Extensible Applications

This section demonstrates the practical application of AS-XAI. The first application concerns interpreting the model's predictions for OOD categories (Section 3.4.1). The second application explores providing auxiliary explanations for difficult-to-distinguish species (Section 3.4.2). The third application evaluates the advantages of our approach in providing trustworthy explanations (Section 3.4.3).

### 3.4.1. Application: Interpreting Out-of-Distribution Categories

In recent years, out-of-distribution (OOD)[42] detection has been critical to ensuring the reliability and safety of machine learning systems. Models are typically trained on a limited dataset to achieve generalization. However, investigations on model interpretation for OOD data are still in their early stages. This application aims to demonstrate the proposed AS-XAI shared semantic features among in-distribution categories that resemble those of OOD categories. Importantly, it answers the 'where, what, and why' questions in CNN decision-making from a semantic perspective.

To explore the interpretation of the model on complex samples beyond the existing dataset, we infer some typical hybrid species and unique species with complex appearances. For instance, consider the case of a liger sample with a tiger maternal lineage and a lion paternal lineage. The liger's appearance usually exhibits the following characteristics: its body markings and legs are similar to those of a tiger, while its head details resemble a lion, such as the mane, nose, and eyes. As depicted in **Figure 5a**, VGG-19 identifies a liger as a lion with a probability of 57.4%; whereas, AS-XAI detects significant semantic features according to the image and searches for similar semantics of in-distribution data trained by the model. The red irregular region in the figure represents relatively salient semantic regions. The experiment demonstrates that AS-XAI can successfully excavate the maternal and paternal categories of the liger through similar characteristics of the in-distribution samples. Furthermore, we input an image of a tapir as an OOD category (Figure 5b). Unlike typical hybrid species, the tapir shares similar appearance characteristics with various animals, including an elongated nose, robust skin, and relatively small eyes and ears, despite being a single species. VGG-19 recognizes it as a pig with only 5.7% accuracy, which is nearly at random guessing levels. In our work, AS-XAI first identifies all important semantic regions of the tapir and calculates the probability that these semantics occur in the semantic space of in-sample categories. As shown in Figure 5b, the proposed AS-XAI analysis shows that the ears, head, nose, eyes, and legs of a tapir are similar to the ears of a donkey, the head of a pig, the nose of an elephant, and the eyes and legs of a rhinoceros, respectively $PCS_s(z) > 0.6$, where $s$ represents the above four semantics. Therefore, even for intricate species that are challenging to identify, AS-XAI can provide detailed semantic explanations by identifying common traits and distinctions across species.

From the experiments, AS-XAI can model the relationship between categories in the feature space, exploring the similarities and differences of common semantic features between different categories. This method explains significant semantic attributes in OOD data using existing datasets, which makes up for the inherent shortcomings of CNN's lack of confidence due to limited data generalization. This unique interpretation approach enables the model to capture the important features of OOD categories and embodies the interpretation process more intuitively by combining semantic probability and visualization.

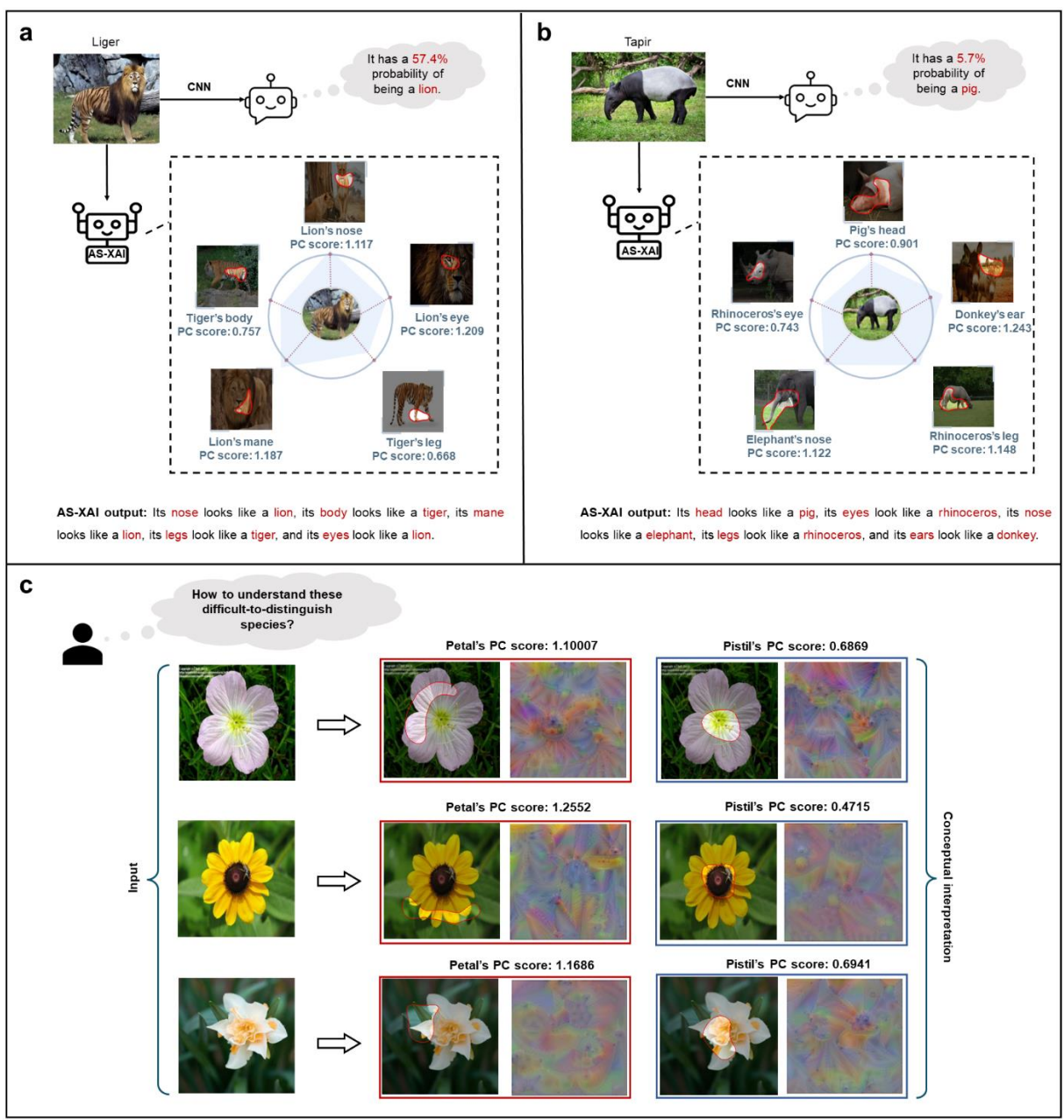

**Figure 5.** Practical applications of AS-XAI. a) Explaining OOD hybrid species using shared semantic features of in-distribution categories. b) Explaining OOD unique species with complex appearance. c) AS-XAI assisted interpretation of species that are difficult for humans to distinguish, left: input images, right: interpret the importance of distinctive plant attributes in category distinction.

*3.4.2. Application: Auxiliary Explanations for Difficult-to-Distinguish Species*

The Earth is rich in species, some of which are so similar in characteristics that it is difficult for non-specialists to position an entry point for identifying these species. Therefore, rapid and accurate species identification has become an important challenge in biodiversity research. In previous interpretable investigations,[43] animal classification was usually recognized, and the methods were not migrated to plant recognition. In addition, unlike animals, in which the distribution of important features is more obvious, the important features of plants are often irregular. Using a rectangular mask as the background will bring a large amount of superfluous feature information, which greatly reduces the usability of the interpretable methods on plants that are difficult for humans to understand.

In this study, AS-XAI utilizes a self-supervised approach to extract semantically important irregular regions combined with the importance of sensitive semantic concepts to provide auxiliary explanations for species that are challenging for humans to distinguish. We conduct sufficient experiments to reflect the better generalization of the interpretation method in different species recognition, the detailed process of which is provided in Supplementary Information S.3. Here, we take Black-eyed Susan as an example. Most people may have heard the names of these flowers, but recognizing them as effortlessly as professionals can be relatively difficult. As shown in Figure 5c, the irregular semantic regions of the species through self-supervision are visualized vividly, so we determine that the semantic focus of CNN to understand the species is petals and pistils. Then, the PCS of the petal is 1.2552, and the PCS of the pistil is 0.4715 through further calculation of the sensitive semantic features. Therefore, from the perspective of CNN's understanding, the semantic importance of Black-eyed Susan's petals is higher than that of pistils.

The above experiments show that AS-XAI can accurately display the semantic cognition of CNN for different species and provide a reasonable basis for human beings to understand these difficult-to-distinguish species. Simultaneously, this interpretable method can be well extended to data other than animals, which plays a significant role in biodiversity research.

*3.4.3. Application: Trustworthy Interpretation*

In CNN, classification is generally accomplished by outputting the probability of each label. However, this method fails to provide sufficiently reliable information and may even yield misleading results due to its lack of specificity in judgment. As CNN primarily emphasizes image texture recognition,[22] explaining the black-box model's decisions using SoftMax probabilities is inadequate for images with similar textures. This application illustrates that the proposed AS-XAI can provide globally and locally trustworthy interpretations by integrating image semantic probabilities with its color and structural similarities. It corrects CNN's overconfident interpretation of confusing perspectives in images, addressing the issue of semantic ambiguity in fine-grained classification tasks.

On the counterfactual image task of cats, we observed that VGG-19 tended to make overconfident judgments regarding images with confusing perspectives. AS-XAI effectively corrected these misinterpretations from a semantic standpoint, providing a comprehensive and trustworthy interpretation. For instance, as shown in **Figure 6a**, for an image that does not capture the entire body of a cat, which might be mistaken for other animals with similar body shapes (dogs, leopards, tigers) or a plush doll by humans, VGG-19 still draws an overly confident conclusion with a 100% probability of being a cat. In this work, AS-XAI combines semantic possibility and semantic importance. It is evident that the PCS for interpreting this image as a cat and a dog is 0.09 and 0.07, respectively. Based on the semantic decision rules that we provided (see Extended Data Table 2), AS-XAI judges that the image has confusing semantic features from a semantic perspective, and outputs "I am not sure whether this is a cat or a dog". Furthermore, we provide an image of a cat from a confused viewpoint (Figure 6b), in which the VGG-19 predicts it to be a cat with a confidence of 98.6%, but humans confusingly distinguish it as a carpet. In this study, AS-XAI utilizes subjective important semantic visualization and trusted interpretation to effectively identify salient semantic concepts in the test image that influence CNN decisions. Figure 6b reveals a higher similarity between the image's legs and the common semantic space for cat legs in the dataset.

For the classification task using dog and cat data as an example, AS-XAI solves the problem of fine-grained semantic ambiguity and obfuscation by integrating global and local interpretation. Figure 6c shows that the VGG-19 correctly distinguish the input picture as a dog with 100% probability. The bubble ring diagram shows that VGG-19 can easily semantically confuse dog ears with cat ears. The dog's eyes, legs, and nose are relatively more important for VGG-19 decision-making. The structural similarity histogram further confirms that the pixel values of all patches in this image are more similar to the dog image in the dataset. Consequently, the output of AS-XAI's interpretation is: "I am sure it is a dog mainly because it has a vivid nose, which is a dog's nose obviously. Meanwhile, it has eyes, which are something like a dog's eyes. In addition, it seems to have confused ears. The dog shows a higher semantic similarity score." For fine-grained multi-classification problems, AS-XAI can also accurately distinguish the differences between classes to avoid semantic confusion between classes (Figure 6d). VGG-19 only provided a 98% probability of identifying a Russian blue cat, while the proposed AS-XAI interprets the image as a cat through the bubble ring diagram. Moreover, utilizing the structural similarity histogram, AS-XAI can more specifically explain that each part of this cat image is closer to the Russian blue cat.

From the above experimental results on the counterfactual and categorization tasks, it is evident that the images' semantic space and pixel space play complementary roles in the interpretability of the categorization task. In counterfactual tasks, semantic space interpretation corrects CNN's overconfidence in the semantic obfuscation perspective. On multi-classification tasks, pixel-space judgments will compensate for the limitations of semantic space for fine-grained category judgments. Therefore, AS-XAI can provide a more comprehensive explanation of global decision-making in CNNs.

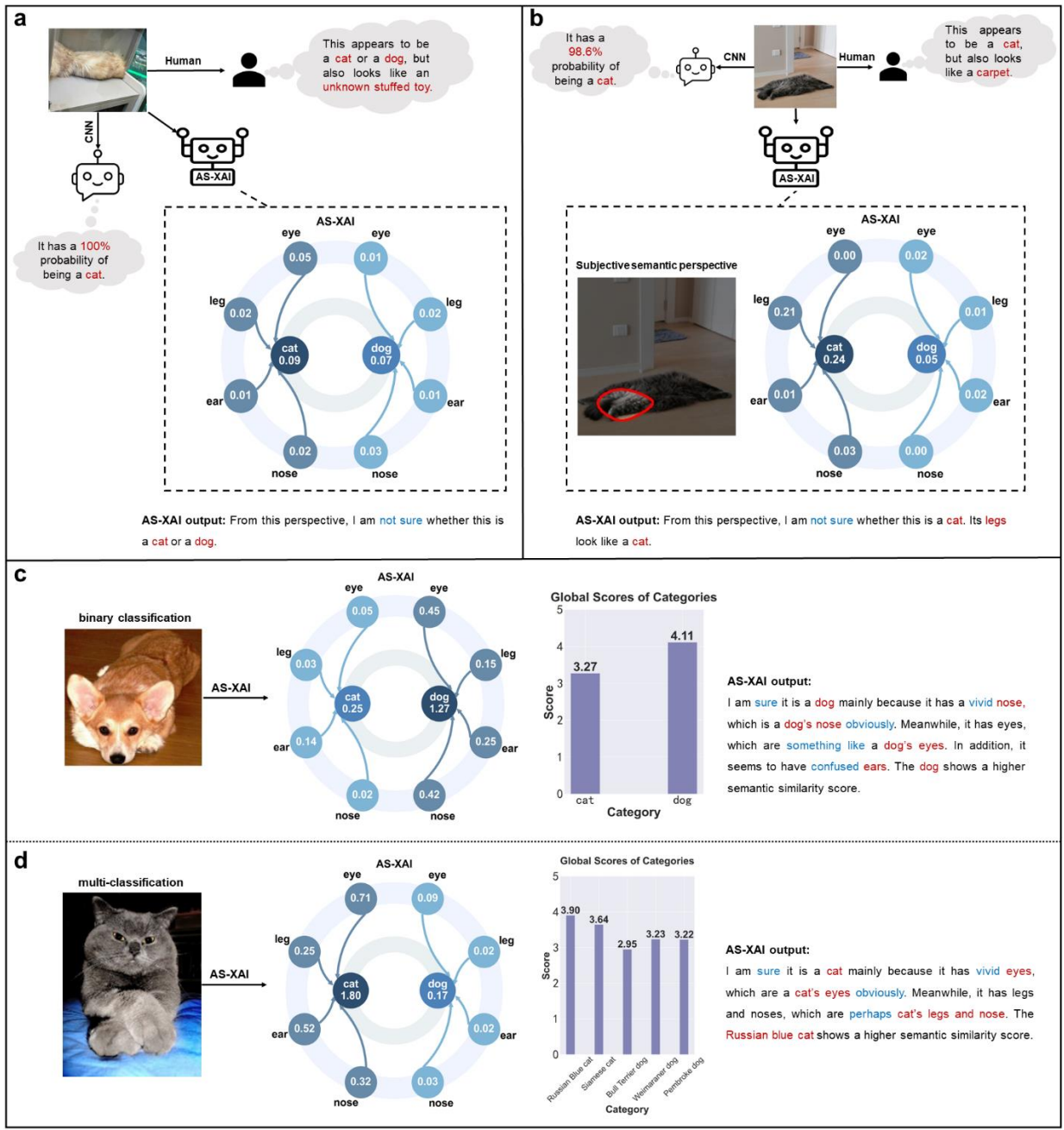

**Figure 6.** Trustworthy interpretation of AS-XAI. a) Human (right), Proto-CNN (bottom), and AS-XAI (bottom right) interpretation of the image when the entire body of the cat is not captured. b) Global semantic interpretation of a cat counterfactual image by AS-XAI. c) Interpretation of AS-XAI for binary classifications of dog and cat datasets. d) Interpretation of AS-XAI for multiple classifications of dog and cat datasets.

### 3.5. Orthogonality Ablation Experiment

Since the extracted semantics do not have the label of ground truth, it is difficult to measure the advantages and disadvantages of masks by common indicators, such as recall, mean average precision, etc.[19] In addition, the diversity of samples will greatly affect the learning and prediction ability of algorithms. Although previous studies attempted to add meaningful semantic feature categories by manual annotation artificially,[34] the independence of semantics cannot be guaranteed. Therefore, this application conducted ablation experiments comparing the common semantic traits from orthogonal and non-orthogonal self-supervised AS-XAI to weakly orthogonal non-self-supervised AS-XAI. The results of the experiments verify that strictly orthogonal self-supervised AS-XAI can obtain more accurate global semantic interpretations.

We randomly selected $N_s$ cat and dog datasets, and extracted important semantic concepts for orthogonality calculation. **Figure 7a,b** show the orthogonal relationship between the semantic concepts extracted by the self-supervised AS-XAI and the non-self-supervised AS-XAI in the cat and dog datasets. In an orthogonal matrix, the closer the value is to 0 (the color bar is white), the stronger is the orthogonality between properties, and vice versa. From the orthogonal matrix, it is evident that the semantic features of self-supervised AS-XAI have much better orthogonality. To evaluate the direct impact of orthogonality on semantically interpretable results, we conducted a series of ablation experiments comparing the weakly orthogonal non-self-supervised AS-XAI and self-supervised AS-XAI with and without orthogonality constraints. The results, shown in Figure 7c, indicate that the visualization results of self-supervised AS-XAI without adding orthogonality loss exhibit complex features of intertwined or mixed semantics, failing to obtain distinct semantic concepts. For example, since most cats have concentrated facial semantics, local facial entanglement appears in the common semantic space of their eyes, which is reflected in the second row of the first column in Figure 7c. Visualizations of weakly orthogonal non-self-supervised AS-XAI show large-scale semantic confusion (see the last row of Figure 7c), such as mixed leg features in the common traits of the cat's eyes. In contrast, the extraction of clearer and purer semantic concepts is facilitated by strongly orthogonal self-supervised AS-XAI, enabling the discovery of a more diverse array of semantic features.

In previous interpretation work[34] based on semantic space, semantic concepts are usually defined by humans, so it is revealed that a neuron is sensitive to multiple semantic regions simultaneously. However, this study employs semantic orthogonality to verify the rationality of previous research results. The findings also demonstrate that AS-XAI can avoid label bias caused by manual extraction of semantic features through self-supervision, which facilitates distinguishing the differences between semantic concepts learned by the model and semantic concepts recognized by humans.

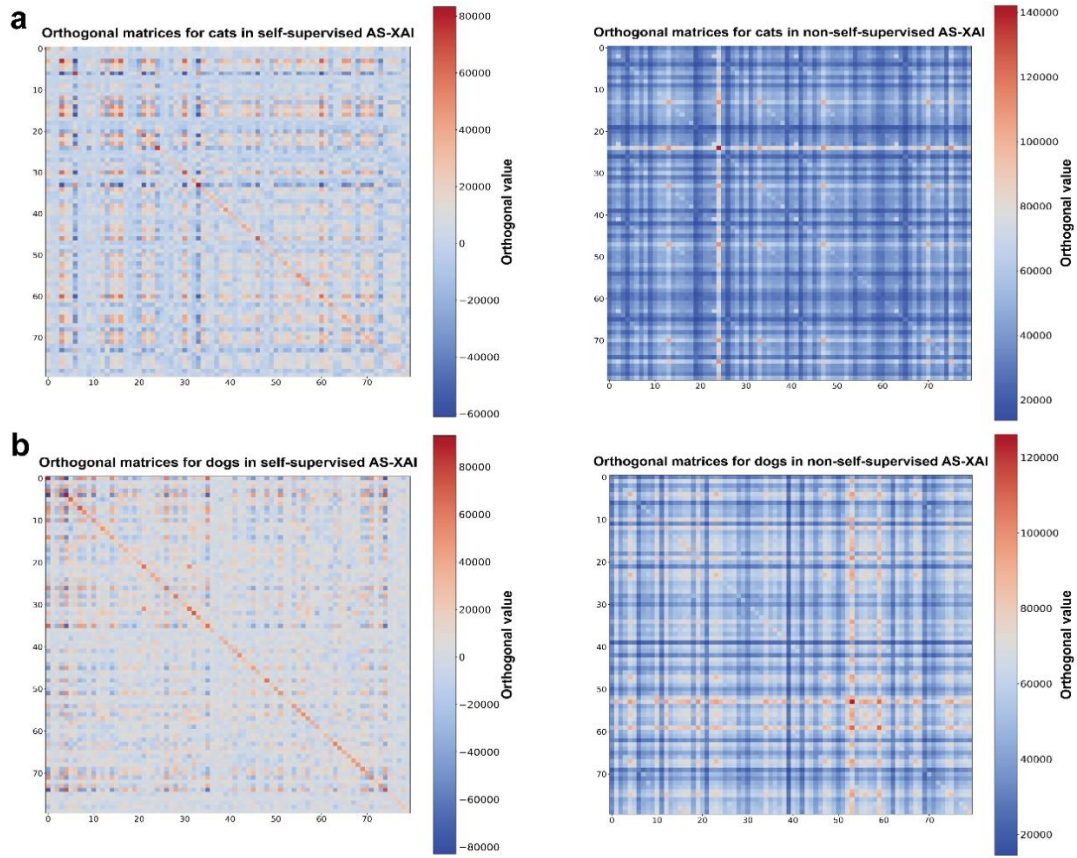

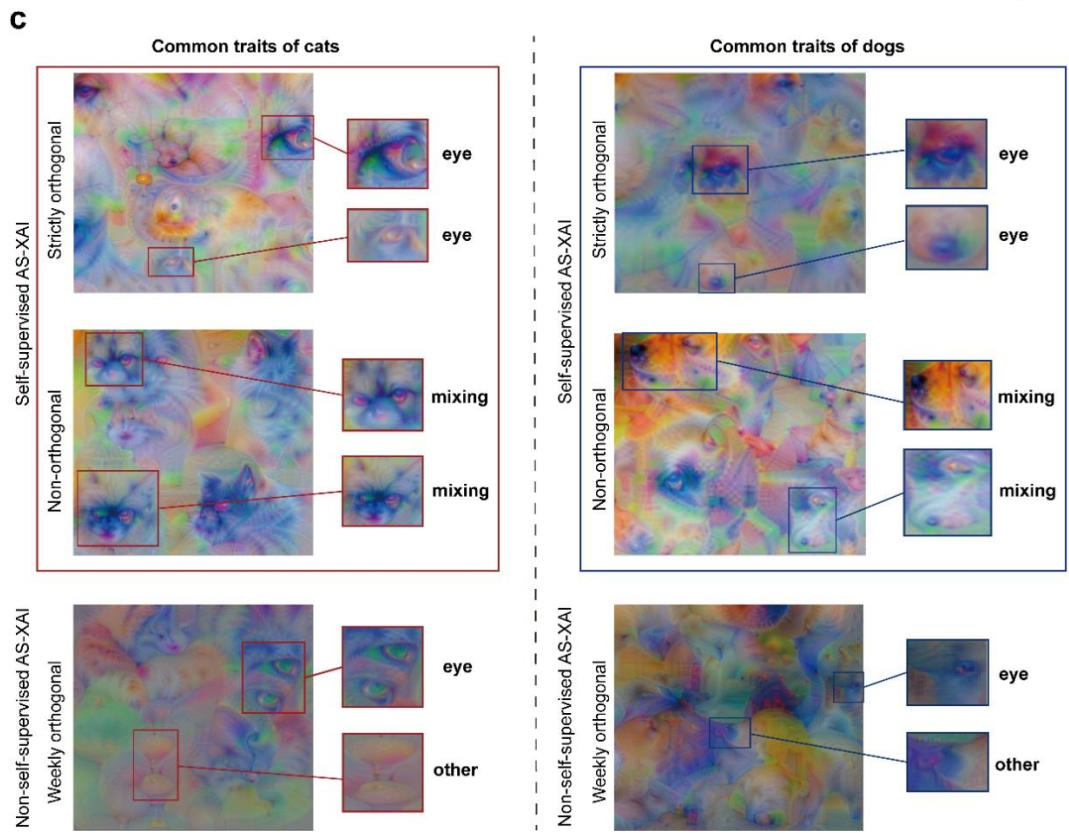

**Figure 7.** Orthogonality ablation experiment in interpretable semantic spaces. Comparison of the orthogonality of semantic concepts ($N_s = 80$) extracted from the cat dataset (a) and the dog dataset (b) by self-supervised AS-XAI and non-self-supervised AS-XAI. c) Visualization of the subjective results of common traits extracted by different methods. Comparison of the subjective results of eye commonality features extracted in self-supervised AS-XAI and non-self-supervised AS-XAI for the cat dataset (the first column) and the dog dataset (the second column). The subjective results of self-supervised AS-XAI with and without orthogonal loss are compared in the box.

## 4. Discussion

In this work, we introduce AS-XAI, a self-supervised semantic explainable artificial intelligence that identifies the relevant parts of input images to common semantic concepts in the training data. AS-XAI combines local pixel and global semantic perspectives, effectively implementing fine-grained decision interpretation of CNNs, considerably extending the state-of-the-art. Furthermore, AS-XAI initially realizes the trade-off among interpretability assessment criteria (reliability, causality, and usability). From the perspective of reliability, this interpretation strategy aims to make the black-box model transparent through self-supervised feature extraction in transparent embedding space and exploration of common semantic space with row-centered PCA, thus ensuring that the interpretation logic of the proposed AS-XAI strictly adheres to the decision-making mechanism of CNN. From the perspective of causality, the proposed AS-XAI method evaluates the importance of semantic concepts extracted by the model based on the decomposition invariance of the semantic features in the filter. Through combining sensitivity analysis and semantic concept visualization, the contributions from different layers of filters can be accurately attributed and explicitly located to specific semantic concepts by AS-XAI. This analysis provides a clear understanding of the internal decision logic of the black-box model and the cognitive basis for judging the samples. From the usability perspective, this interpretation method extracts important semantic features in samples through self-supervision and carries out lightweight black-box interpretation. AS-XAI can efficiently provide semantic interpretation for CNN in the absence of human interference and without additional computational cost. It has fine-grained and scalable practical usage, including trusted interpretation and semantic retrieval, interpreting OOD categories, and auxiliary interpretation of species that are difficult for humans to understand.

The experiments have qualitatively and quantitatively demonstrated that the self-supervised AS-XAI can systematically assist humans to understand the decision-making process of the black-box model on samples, which has potential value for in-distribution, OOD datasets, and end-to-end trained models. The results validate the effectiveness of AS-XAI in achieving global semantic interpretation by extracting common traits from the data. In contrast, traditional single-sample semantic interpretation is often difficult for humans to interpret and comprehend semantic concepts. Furthermore, the proposed AS-XAI avoids subjectivity and bias from manual labeling. Orthogonal ablation experiments demonstrate that semantic features extracted strictly orthogonally by AS-XAI are more targeted, significantly improving trustworthy interpretation. In addition, AS-XAI also proves variations in CNN sensitivity to diverse pixel perception domains during feature extraction. Specifically, CNN prioritized the shape and contrast perception of sample semantics, which were inconsistent with the pixel perception preference of the human visual system, as described in Supplementary Information S.1. Finally, the proposed AS-XAI is proven to be effective in providing a fine-grained and extensible interpretation framework for various real-world tasks.

Overall, AS-XAI extracts the comprehensible semantic space in a self-supervised manner, which allows us to visualize the network's semantic understanding of the target concepts without compromising the model performance. Additionally, it demonstrates the difference between the network's decision-making and human cognition. The proposed AS-XAI aims to utilize neural network explanations as a means of scientific reasoning, which can provide heuristic explanations for more complex neural network architectures. Future work could involve the development of an adversarial defense mechanism to enhance network robustness based on the sensitivity differences of CNNs to semantic space. Furthermore, we will explore how the network comprehends the color and texture semantics of target samples, aiming to better elucidate the black-box model from a semantic perspective. We believe that the semantic transparency interpretation of AS-XAI to black-box models will advance the applicability of XAI in novel or challenging domains.

**Author contributions**

C.S. implemented the workflow, developed the code, analyzed the data, and wrote the manuscript; H.X., Y.C., and D.Z. conceived the project and designed the research; Y.C. and D.Z. supervised the whole project.

**Code availability**

We use PyTorch (https://pytorch.org) for deep model training, and the code for replicating our experiments is available on GitHub (https://github.com/qi657/AS-XAI).

**Conflicts of interest**

The authors declare no conflicts of interest.

**Acknowledgements**

The authors express sincere gratitude to Xinyue Liu, Longfeng Nie, and Zhengguang Liu for valuable discussions. This work was supported and partially funded by: The National Center for Applied Mathematics Shenzhen (NCAMS); The Shenzhen Key Laboratory of Natural Gas Hydrates (Grant No. ZDSYS20200421111201738); The SUSTech – Qingdao New Energy Technology Research Institute; The National Natural Science Foundation of China (Grant No. 62106116); and The Major Key Project of PCL (Grant No. PCL2023A09).

# Extended Data

**Extended Data Table 1.** Typical interpretable methodology review.

| Type | Method | Descriptions | Limitation |
|---|---|---|---|
| Local interpretation | Perturbation-based forward propagation[1] | Perturb a single test sample input or intermediate layer, and calculate the effect on the output. | Repeated perturbations lead to high computational costs, and their interpretation results are intuitively difficult to understand. |
| | Perturbation-based backpropagation method[2] | Explain model predictions by propagating importance. | Different transfer rules will lead to differences in interpretation results. |
| | Saliency-based methods[3] | Calculate the relative importance of features by integrating. | Categories are not considered, thus resulting in very similar interpretations for multi-classification tasks. |
| | Class activation mapping(CAM)[4] | Model interpretability via Shapley values. | Local interpretation of specific layers of the model, and coarse regional localization. |
| | Interpretable representations learning[5] | Learning to interpret the decision-making process by obtaining clear semantic representations of the network training process. | The latent concept space may be related to many different predefined concepts, but the causal relationship is unclear. |
| Global interpretation | Concept relevance propagation (CRP)[6] | Computational models learn concept importance flows through conditional backpropagation. | Single sample interpretation, and high computational cost. |
| | Semantic interpretation[7] | Constructing semantic spaces for group data and mapping to human-understandable concepts for interpretation. | Relying on definitions of semantic knowledge and lack of reasoning about the importance of features, and the scope of practical applications is limited. |

**Extended Data Table 2.** Rules for generating the explanation by AS-XAI.

| | $\Delta_{max}PCS_s(z) < 0.1$ | $0.1 < \Delta_{max}PCS_s(z) < 0.35$ | $0.35 < \Delta_{max}PCS_s(z) < 0.5$ | $\Delta_{max}PCS_s(z) > 0.5$ |
|---|---|---|---|---|
| | $C_d{}^3 PCSs(z) < 0.2$ | $0.2 < C_d{}^l PCS_s(z) < 0.5$ | $0.2 < C_d{}^2 PCS_s(z) < 0.5$ | $C_d{}^l PCS_s(z) > 0.5$ |
| **Assessment** | From this perspective, I am not sure whether this is a cat or a dog. | From this perspective, I am not sure whether this is a dog/cat mainly because | It is probably a dog/cat mainly because | I am sure it is a cat/dog mainly because |
| **Explanation** | $\Delta PCS(z) < 0.1$ | $0.1 < \Delta PCS(z) < 0.5$ | $0.1 < \Delta PCS(z) < 0.35$ | $0.35 < \Delta PCS(z) < 0.5$ | $\Delta PCS(z) > 0.5$ |
| $PCS_{max}(z) > 0.5$ | Position: vivid, Semanteme: confusing | Position: vivid, Semanteme: something like | Position: vivid, Semanteme: perhaps | Position: vivid, Semanteme: something like | Position: vivid, Semanteme: obviously |
| $0.1 < PCS_{max}(z) < 0.5$ | None | Position: be, Semanteme: something like | Position: be, Semanteme: perhaps | Position: be, Semanteme: something like | Position: be, Semanteme: obviously |

**Supplementary Information for**

**AS-XAI: Self-supervised Automatic Semantic Interpretation for CNN**

Changqi Sun[1,2], Hao Xu[3], Yuntian Chen[4, *], and Dongxiao Zhang[2, 4, *]

[1] School of Environmental Science and Engineering, Southern University of Science and Technology, Shenzhen 518055, Guangdong, P. R. China

[2] Department of Mathematics and Theories, Peng Cheng Laboratory, Shenzhen 518000, Guangdong, P. R. China

[3] BIC-ESAT, ERE, and SKLTCS, College of Engineering, Peking University, Beijing 100871, P. R. China

[4] Ningbo Institute of Digital Twin, Eastern Institute of Technology, Ningbo, Zhejiang 315200, P. R. China

[*] Corresponding author.

**The PDF File Includes:**

**S.1:** Supplementary information for exploration of image pixel space.

**S.2:** Supplementary information for auto-extracting common traits.

**S.3:** Supplementary information for auxiliary explanations for difficult-to-distinguish species.

**S.4:** Supplementary information for high-rank sensitive semantic spaces.

**Supplementary Figure 1.** Image modifications for corresponding visual characteristics applied to an exemplar image.

**Supplementary Figure 2.** Box plots of local importance scores across the training sets for each visual semantic concept.

**Supplementary Figure 3.** Comparison of common traits between AS-XAI and S-XAI for dogs.

**Supplementary Figure 4.** Comparison of common traits between AS-XAI and S-XAI for cats.

**Supplementary Figure 5.** Auxiliary explanations for difficult-to-distinguish species.

**Supplementary Figure 6.** Semantic concepts extracted from different datasets at various filter layers.

**S.1: Supplementary Information for Exploration of Image Pixel Space**

As the perceptual domains of humans and CNNs have some differences, human perception of things is biased towards shape, color, size, etc. Since previous work has not considered the combined factors in calculating structural similarity, which makes the local interpretation of the results too one-sided, in this work, we verify the sensitivity of CNN to perturbations in different perceptual domains (i.e., luminance, contrast, hue, saturation, shape, and details) of images. The dataset is first augmented with different aspects of the perceptual domain. Then, the resulting semantic masks are computed by cosine similarity with the original and augmented data to distinguish which information the model pays more attention to different categories of images at a particular semantic level. The data enhancement method that we use is to directly call the ColorJitter function in PyTorch to set the contrast to 0.45, brightness to 0.8, saturation to 0.7, and hue to 0.1, and use the fastNlMeanDenoisingColored function in OpenCV to perform the texture modification of the image, in which the texture modification intensity is set to 4. The texture modification intensity is set to 0.1 for the shapes. The intensity of modification is set to 4. For the shape, the roll function in Numpy is used to perform wheel fluctuation operations on the $H$ and $W$ of the image to make its shape distorted. The data-enhanced image can be seen in **Supplementary Figure 1**. Except for the data pre-processing, the same settings were used during the experiment for the model training process. The formulae for calculating the cosine similarity difference between the semantic mask and all patches of the input image are as follows:

$$\mathcal{L}_{\text{aggregation}} = \frac{1}{n} \sum_{i=1}^{n} \min_{j: a_j^{'} \in A(y_i)} \min_{P \in \text{patches}(Z_i)} -\cos\left(\theta(a_j^{'} \cdot P)\right) \tag{1}$$

$$\mathcal{L}_{\text{aggregation}_1} = \frac{1}{n} \sum_{i=1}^{n} \min_{j: a_j^{'} \in A(y_i)} \min_{P \in \text{patches}(Z_i)} -\cos\left(\theta(b_j^{'} \cdot P)\right) \tag{2}$$

$$\delta = \mathcal{L}_{\text{aggregation}} - \mathcal{L}_{\text{aggregation}_1} \tag{3}$$

where $a_j^{'}$ is the semantic mask extracted from the original data; $b_j^{'}$ is the mask extracted after data augmentation; and $\delta$ is the difference item. We randomly sampled 500 samples from each of the five cat and dog data types, and compared the three semantics (ear, eye, leg) with the original image in terms of hue, brightness, contrast, saturation, shape, and texture. The large difference in similarity indicates that CNN is more sensitive to this perceptual domain. As shown in **Supplementary Figure 2**, we found that the sensitivity of CNN to changes in the perceptual domain of the same semantics under the five categories is the same, and the sensitivity to changes in the perceptual domain of different semantics is different. For example, after adjusting different perception domains for ear semantics, the significant sensitivities affecting CNN are shape, contrast, hue, and texture from high to low. The significant sensitivity of eyes under different categories under different influencing factors, from high to low, is shape, contrast, hue, and texture. The significant sensitivity of legs under different categories under different influencing factors, from high to low, is shape, contrast, hue, and texture. Based on the above experimental conclusions, we found that CNN similarity calculation is similar to human similarity reasoning, and human perception considers the influence of various complexity factors. To this end, we convert the image from the RGB domain to the HSV domain before CNN calculates the structural similarity of the semantic mask and different patches of the image, covering hue, saturation, and brightness. In the image HSV color space, we calculate the spatial structural similarity so that CNN has a global consideration of the color and structure space of the image.

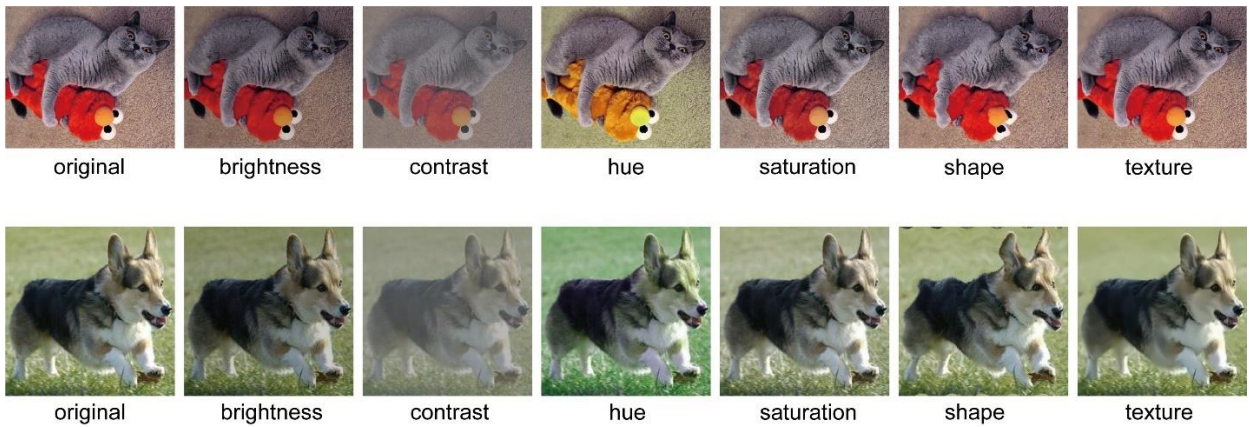

**Supplementary Figure 1.** The image modifications for corresponding visual characteristics applied to an exemplar image.

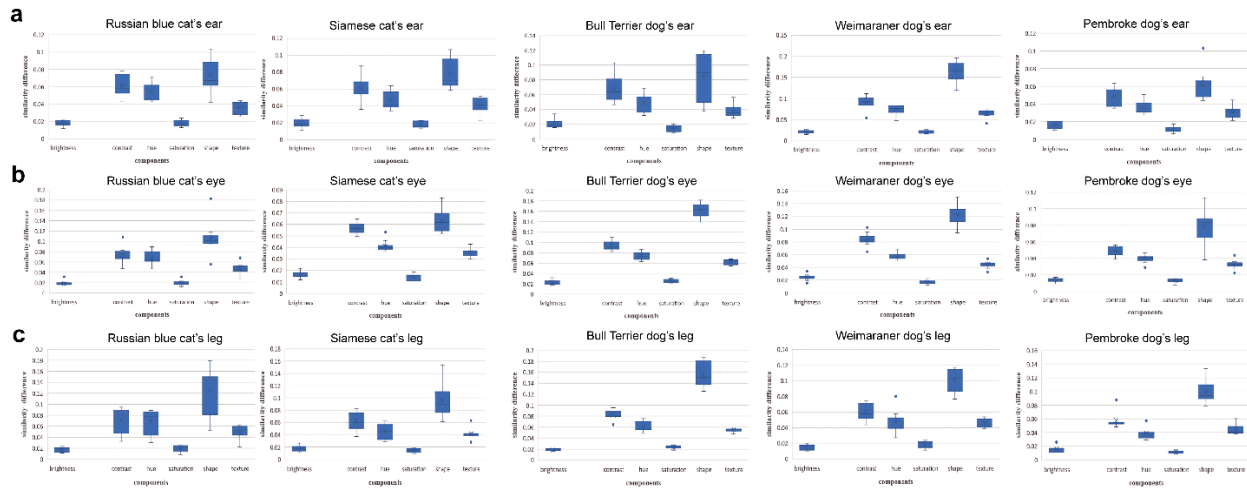

**Supplementary Figure 2.** Box plots of local importance scores across the training sets for each visual semantic concept. Five categories of cats and dogs, a) the sensitivity of ear semantics in different perceptual domains; b) the sensitivity of eye semantics in different perceptual domains; c) the sensitivity of leg semantics in different perceptual domains.

## S.2: Supplementary Information for Auto-extracting Common Traits

In this work, we randomly selected 400 dog samples and 400 cat samples, respectively, from the dataset. Comparison is performed of the visualization results of different semantic spaces (including eyes, ears, nose, and legs) manually extracted by AS-XAI through self-supervised extraction and S-XAI, such as **Supplementary Figure 3** and **Supplementary Figure 4**. Through subjective comparison, it was found that the four semantic common spaces of dogs automatically extracted by AS-XAI reflect strong and targeted features. However, S-XAI is in stark contrast to this method from the subjective point of view, such as the dog's ear. AS-XAI extracts the pure eye common semantic space in the second column of Supplementary Figure 3a,b. Still, S-XAI has a semantic aliasing phenomenon, and the nose appears under the semantic space of the ear, eyes, and legs all semantically. Similarly, in Supplementary Figure 4, the common traits space of the cat's eyes in AS-XAI reflects strong and concentrated eye features, while the common traits space of the eyes in S-XAI is mixed with the leg features. Meanwhile, the common traits space of the legs in S-XAI is mixed with distinct eye features, and this semantic confusion also appears obviously in the common traits space of the ears and nose. From an objective perspective, we use a quantile-quantile (q-q) plot, which is a graphical technique for determining if the given distribution is consistent with a normal distribution. From the q-q plots, the $R^2$ of AS-XAI is above 0.98. It is significantly better than S-XAI, proving that the semantic concept defined by humans is prone to subjectivity and bias. The semantic information learned by the self-supervision of the network can better reflect the understanding mechanism of the network, which also enhances the interpretability of CNN's effectiveness.

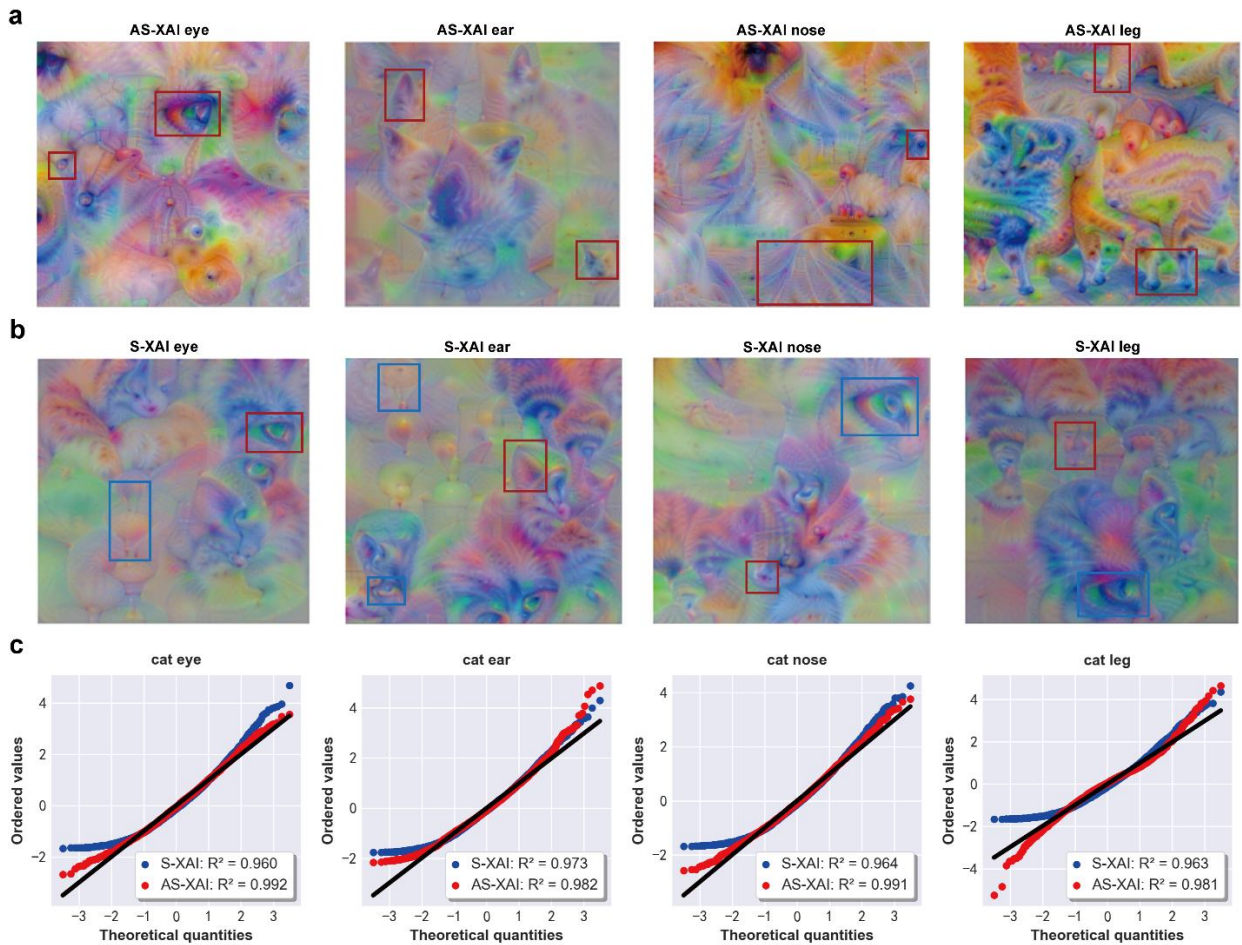

**Supplementary Figure 3.** The figure of common traits between AS-XAI and S-XAI for dogs. a) Visualization of the four semantic spaces of dogs in the AS-XAI method in the 1st PC, where the red box is the accurate feature information in the semantic space; b) visualization of the four semantic spaces of dogs in the 1st PC in the S-XAI method, where the red box is the accurate feature information in the semantic

space, and the blue box is the wrongly confused feature information; c) quantile-quantile (q-q) plots of the four semantic spatial distributions of dogs, where the blue dots are the semantic distributions of S-XAI, the red dots are the semantic distributions of AS-XAI, and the black line is the quantile line fitting a normal distribution.

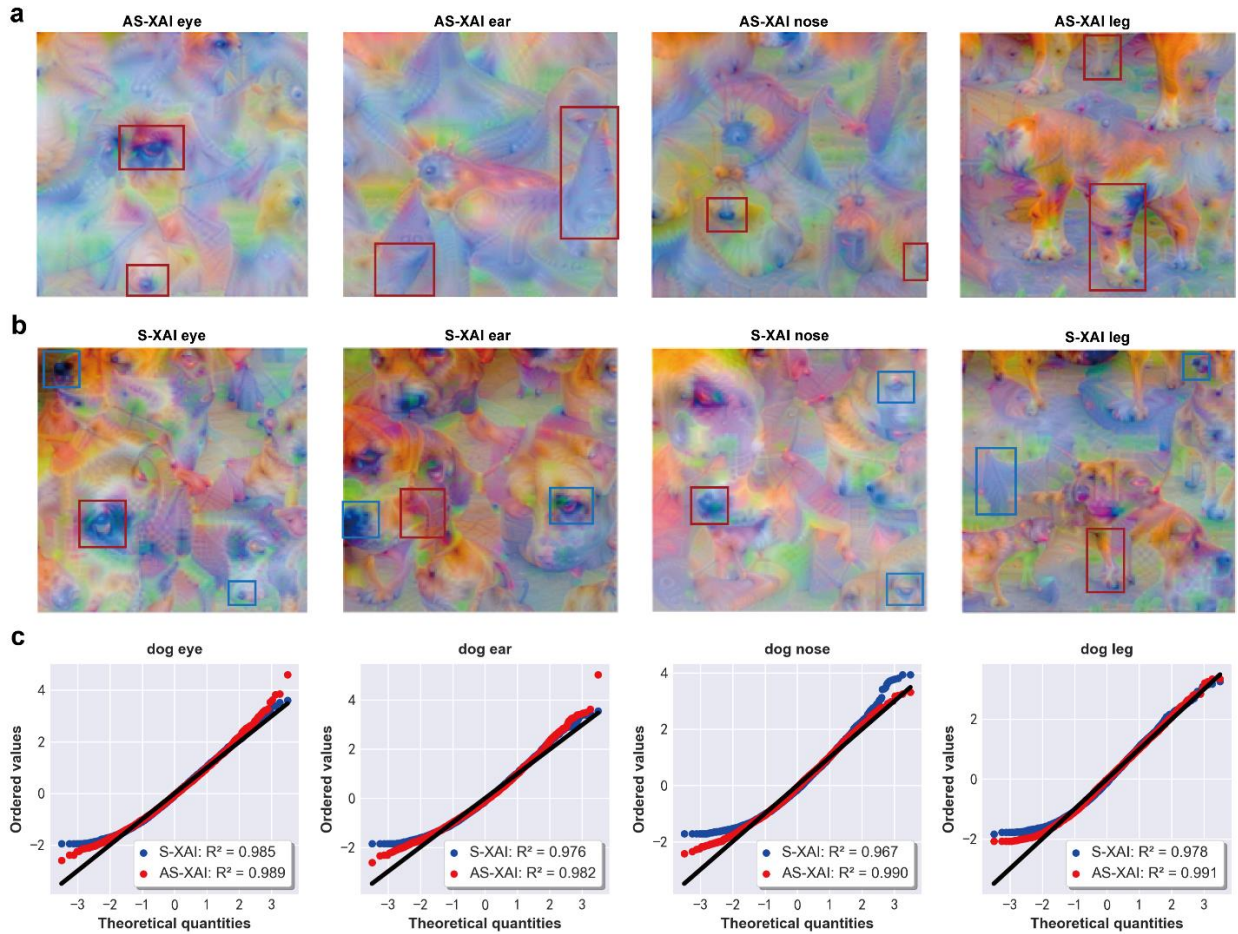

**Supplementary Figure 4.** The figure of common traits between AS-XAI and S-XAI for cats. a) Visualization of the four semantic spatial distributions of cats in the AS-XAI method from left to right for eye, nose, ear, and leg in the $1^{st}$ PC, where the red boxes are the accurate feature information under the semantic spatial distributions; b) visualization of the four semantic spatial distributions of cats in the S-XAI method in the $1^{st}$ PC, where the red boxes are the accurate feature information under the semantic spatial distributions, and the blue boxes denote the erroneous feature information; c) quantile-quantile (q-q) plots

of the four semantic spatial distributions of cats, where the blue dots are the semantic distributions of S-XAI, the red dots are the semantic distributions of AS-XAI, and the black line is the quantization of the fitted normal distribution line.

## S.3: Supplementary Information for Auxiliary Explanations for Difficult-to-distinguish Species

In this work, we use AS-XAI to provide auxiliary interpretation for complex samples that are difficult for humans to understand. For samples with dense semantic scales, irregular semantic masks can present the sensitive areas well and reduce unnecessary semantic redundancy. Through many experiments, we found that this method has an auxiliary effect on identifying simple animals or plants, and has a positive interpretation of animal emotion identification. We interpreted four different plants, as shown in **Supplementary Figure 5a**. The results demonstrated that AS-XAI could accurately locate important semantic regions, such as roots, leaves, and bud centers. Through these features, we clearly understand the important features that affect CNN's decision-making regarding the species, which also shows that AS-XAI can help improve the accuracy and efficiency of species identification. In addition, we also use pictures of dogs with different emotions to validate the method. AS-XAI can still accurately extract the salient features of different emotions without providing any semantic labels. Humans usually observe dogs' angry emotions through their eyes, ears, and mouth. AS-XAI is close to human cognition in emotion recognition. It can also recognize these three semantics (eyes, ears, and mouth) and judge its angry sentiment, as shown in the first and second rows of Supplementary Figure 5b. The fourth row of Supplementary Figure 5b presents a picture of a sad dog. AS-XAI accurately identifies the emotion as "dog sad" based on the distinct semantic features extracted, including the mouth, eyes, and ears.

Based on the experimental results, it was found that AS-XAI self-supervised orthogonal semantic extraction can accurately locate important semantic features of species that are difficult for humans to understand, and can also accurately judge different emotions of animals according to facial emotion semantics. We believe that the better generalization of AS-XAI will enable the method to play an important role in biodiversity research.

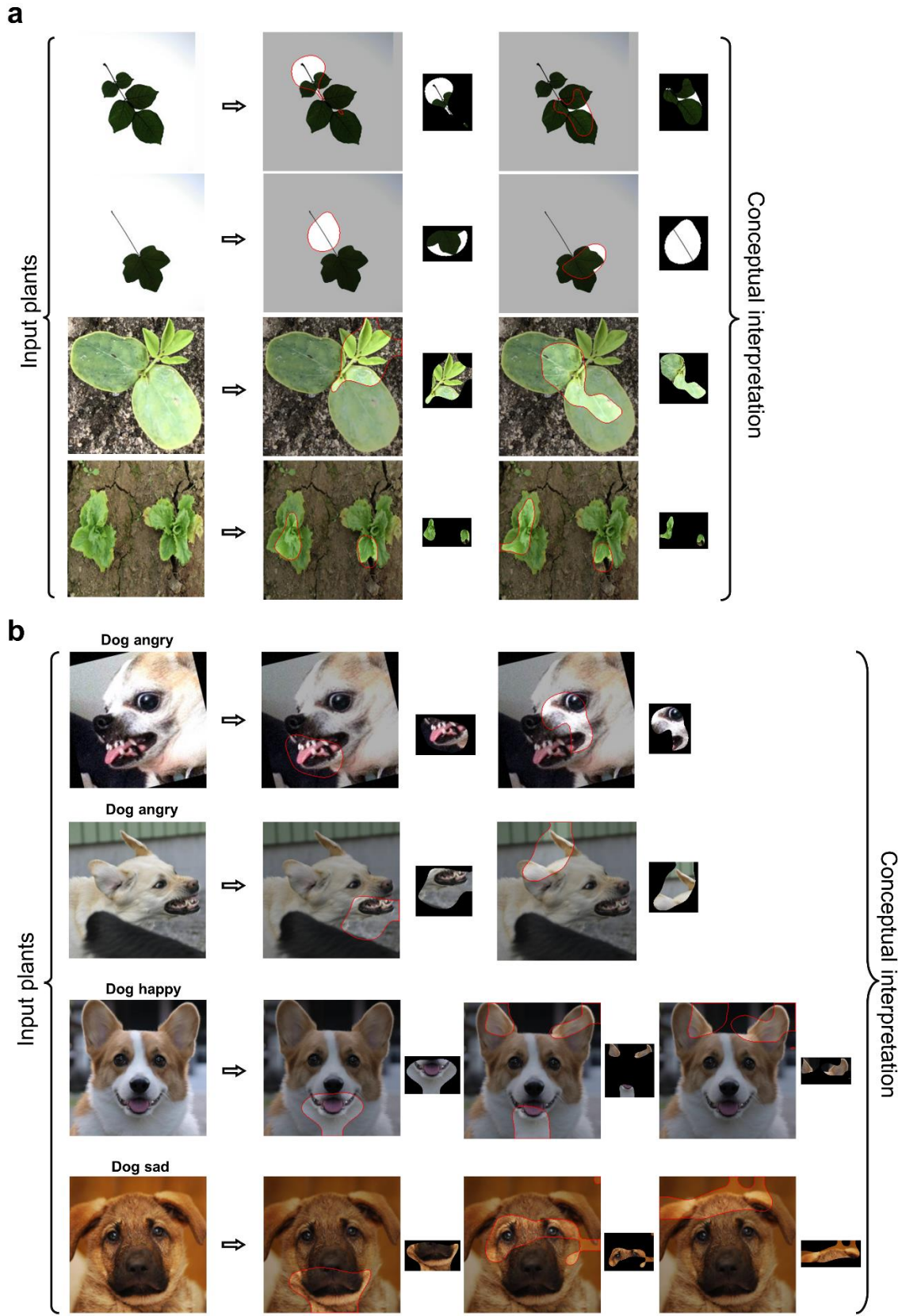

**Supplementary Figure 5.** The figure explanations for difficult-to-distinguish species. a) Auxiliary explanations for plants that are difficult for humans to understand; b) auxiliary explanations for different emotions in dogs.

### S.4: Supplementary Information for High-Rank Sensitive Semantic Spaces

Previous studies have shown that the deeper layers of the network accumulate the shallow features to extract more detailed texture and shape information.[8] In this work, to specify the sensitivity of the convolutional layer to the semantic concepts of the different images of the dataset, we performed an SVD high-rank decomposition of the last layer of the filter for the feature extraction of the network. As shown in **Supplementary Figure 6**, the horizontal coordinate indicates the corresponding order number of the filter, the vertical coordinate indicates the filter rank, and different colours represent different semantic concepts. Based on the semantic high-rank analysis and visualized semantic concepts derived from extensive data, we observe distinctions in the semantic concepts obtained from various image categories within the same filter. Conversely, similarities are apparent in the semantic concepts obtained from identical image categories within the same filter, as depicted in Supplementary Figure 6a,b. Taking the cat and dog data as an example, the same layer id extracts differing semantic concepts across datasets. In addition, we also observed that the 19th and 59th filters can extract sensitive semantic information and have the highest rank values calculated. This further emphasizes the significance of these two filters as crucial layers in the convolution, which are capable of extracting important information. We applied the same method to test sensitive semantic concepts on the primrose dataset. The results showed that certain filters could not extract sensitive semantic information, resulting in a rank of 0 for these filters. Therefore, we conclude that filters may exhibit distinct redundancy patterns when the network processes different datasets due to varying sample distributions, feature distributions, and data noise in different datasets.

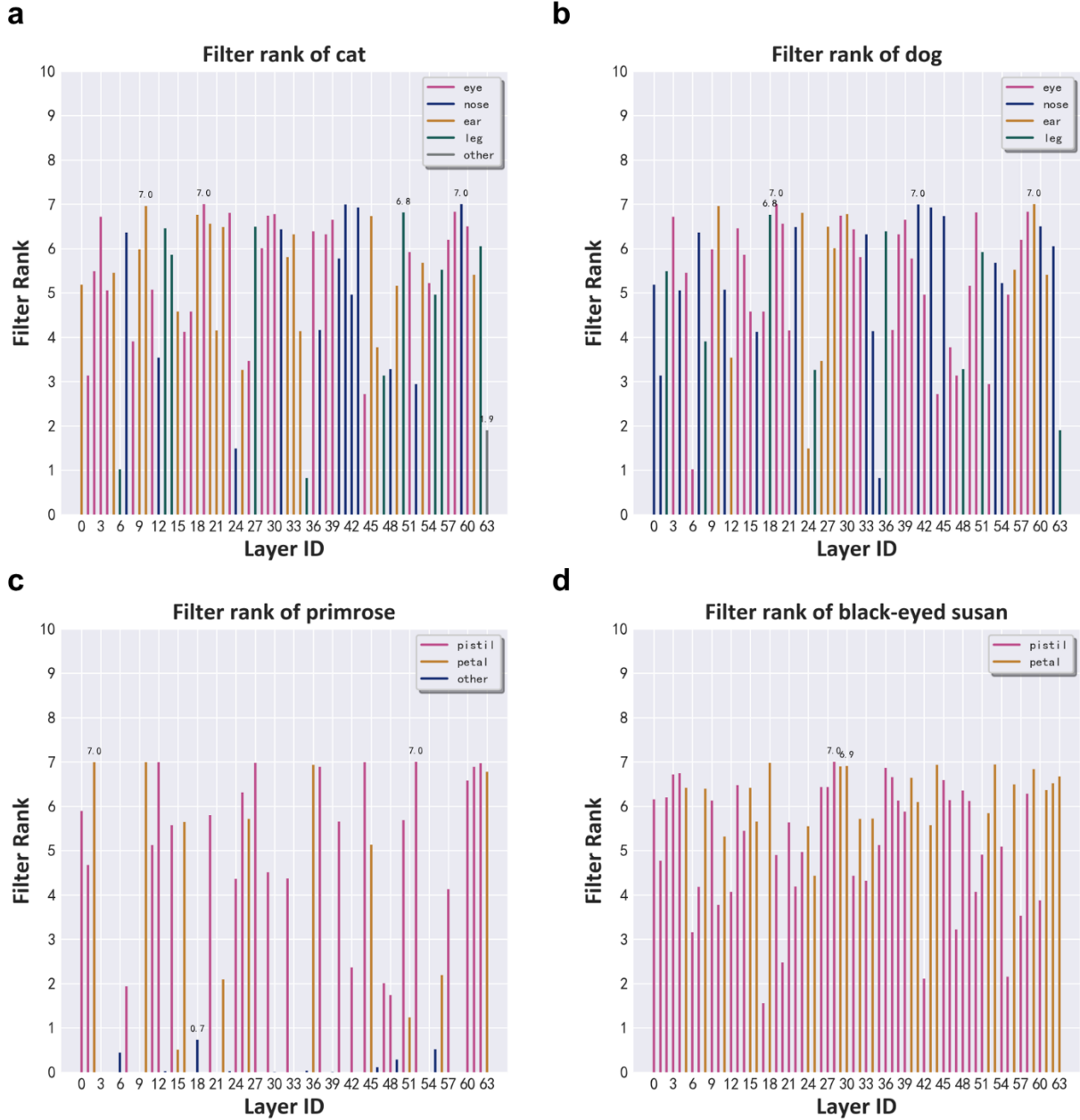

**Supplementary Figure 6.** Semantic concepts extracted from different datasets at various filter layers.